%% file: main.tex
\documentclass[10pt,twocolumn,letterpaper]{article}

\usepackage{wacv}              

\usepackage[accsupp]{axessibility}  
\input{preamble}

\definecolor{wacvblue}{rgb}{0.21,0.49,0.74}
\usepackage[pagebackref,breaklinks,colorlinks,allcolors=wacvblue]{hyperref}
\def\wacvPaperID{591} 
\def\confName{WACV}
\def\confYear{2026}

\title{ViSTA: Visual Storytelling using Multi-modal Adapters \\ for Text-to-Image Diffusion Models}

\author{
\begin{tabular}{ccccc}
Sibo Dong & Ismail Shaheen & Maggie Shen & Rupayan Mallick & Sarah Adel Bargal 
\end{tabular}\\[0.5em]
\begin{tabular}{c}
{\tt\small \{sd1242, ias68, xs196, rupayan.mallick, sarah.bargal\}@georgetown.edu}  \\
Department of Computer Science, Georgetown University, Washington, D.C., USA
\end{tabular}
}

\begin{document}
\twocolumn[{%
\renewcommand\twocolumn[1][]{#1}%
\maketitle
\vspace{-1cm}
\begin{center}
    \centering
    \captionsetup{type=figure}
    \includegraphics[width=0.98\textwidth]{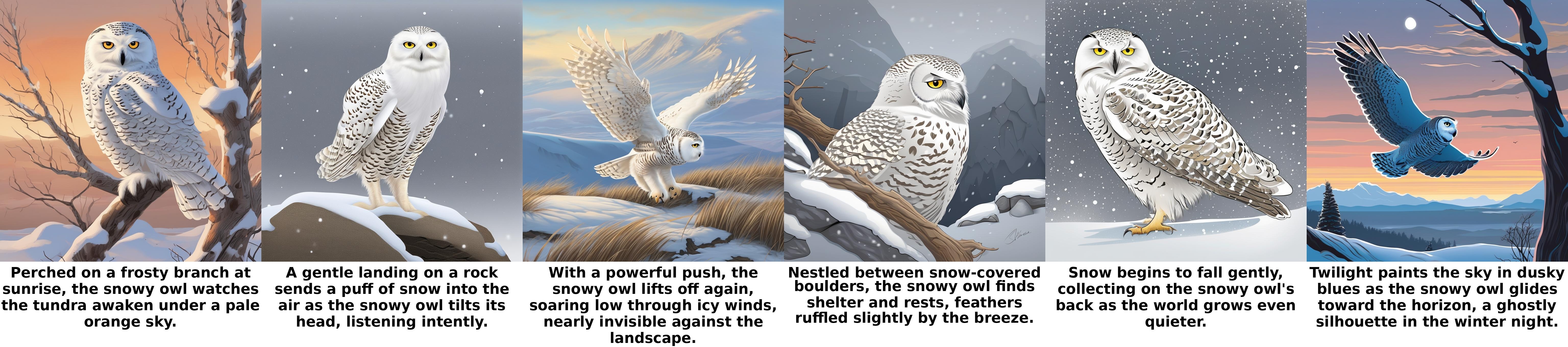}
    \includegraphics[width=0.98\textwidth]{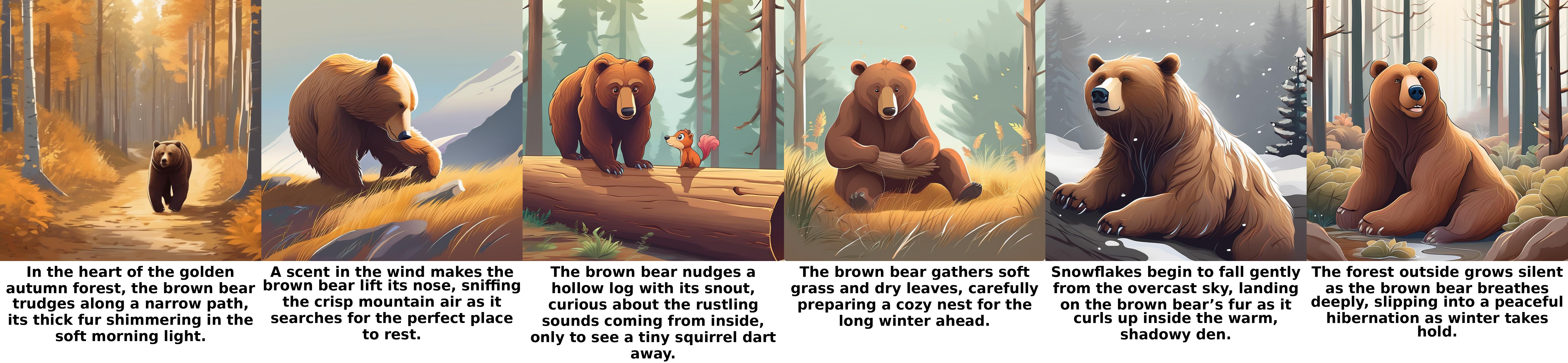}
    \captionof{figure}{ViSTA sample visual storytelling. Given a sequence of narrative story prompts and the first text-image pair as initial references, we generate high-quality story images aligned with their corresponding prompts, maintaining character consistency. The images are generated in an auto-regressive way, where all previous text-image pairs are given, referred to as history text-image pairs. }
  \label{fig:demo}
\end{center}%
\vspace{0.3cm}
}]
\begin{abstract}
Text-to-image diffusion models have achieved remarkable success, yet generating coherent image sequences for visual storytelling remains challenging. A key challenge is effectively leveraging all previous text-image pairs, referred to as history text-image pairs, which provide contextual information for maintaining consistency across frames. Existing auto-regressive methods condition on all past image-text pairs but require extensive training, while training-free subject-specific approaches ensure consistency but lack adaptability to narrative prompts. To address these limitations, we propose a multi-modal history adapter for text-to-image diffusion models, \textbf{ViSTA}. It consists of (1) a multi-modal history fusion module to extract relevant history features and (2) a history adapter to condition the generation on the extracted relevant features. 
We also introduce a salient history selection strategy during inference, where the most salient history text-image pair is selected, improving the quality of the conditioning. Furthermore, we propose to employ a Visual Question Answering-based metric TIFA to assess text-image alignment in visual storytelling, providing a more targeted and interpretable assessment of generated images.
Evaluated on the StorySalon and FlintStonesSV dataset, our proposed ViSTA model is not only consistent across different frames, but also well-aligned with the narrative text descriptions.
\end{abstract}

\section{Introduction}
With the rapid advancement of diffusion models~\cite{DDPM,DPM}, text-to-image generation ~\cite{GLIDE, Dalle2, imagen, sd, sdxl} has achieved remarkable quality and has been widely applied to different tasks. While common text-to-image generation focuses on producing a single high-quality image from a given prompt, visual storytelling presents a more challenging setup by requiring the generation of a sequence of images that maintains consistency in characters, objects, and style across different frames. In visual storytelling, the model takes a sequence of text prompts forming a coherent storyline and generates the corresponding images. Based on whether initial reference images are provided, the task can be further categorized into two settings: story continuation, where an initial reference image is given, and story generation, where no reference image is given. 

Recent works have explored visual storytelling employing diffusion models in an auto-regressive manner \cite{arldm, makeastory, acmvsg, storygen, causalstory, yang2024seedstory}, where, in addition to the current text prompt, all previous text-image pairs, referred to as history text-image pairs, are provided as additional conditions for generating the next image. However, one limitation of these approaches is that not all history contributes equally to the generation process. To address this, several works \cite{acmvsg, causalstory, storygen} propose different methods to reweight the history text-image pairs.
Despite their effectiveness, previous auto-regressive methods share another common limitation. The inclusion of history text and images as additional conditioning signals necessitates modifications to the diffusion model, resulting in the need for extensive training to adapt the diffusion model to these changes. In contrast, our approach processes history context using a novel multi-modal fusion setup, in addition to a lightweight adapter architecture---an efficient strategy demonstrated by IP-Adapter \cite{ip-adapter}---allowing seamless integration of history features without modifying the diffusion model.


Another line of research focuses on subject-consistent generation, which aims to maintain visual consistency across images in a training-free, plug-and-play manner, such as StoryDiffusion \cite{zhou2024storydiffusion}, ConsiStory \cite{tewel_training-free_2024}, and OnePromptOneStory \cite{onepromptonestory}. Unlike auto-regressive methods, which rely on long narrative-style text descriptions, these approaches typically follow a structured prompt format of ``character description + activity.'' Although these methods can ensure character consistency, they have notable limitations: (1) They do not support reference images, meaning that they cannot generate specific, predefined characters. (2) To enforce consistency, the same character prompt is used in all frames, limiting the model’s ability to adapt to varying narrative descriptions or multiple character stories, thus constraining the flexibility of storytelling.

A key characteristic of the visual storytelling task is the presence of history text-image pairs. These history texts and images contain rich contextual information that can effectively guide the generation of subsequent images. For example, prior images and text may include overlapping characters, objects, or implicit references that contribute to maintaining  consistency throughout the story. Furthermore, history text-image pairs naturally align, as the text provides a description of the visual elements present in the images.

In this paper, we tackle the visual storytelling task, where the goal is to generate a sequence of images that are both consistent across frames and well-aligned with narrative text descriptions. 
A central challenge in this task is how to effectively incorporate multi-modal history---previous text-image pairs---into the image generation process. To address this, we propose ViSTA, a lightweight multi-modal history adapter for text-to-image diffusion models. ViSTA enables coherent and text-aligned story generation by leveraging history text-image pairs through a dedicated fusion and conditioning mechanism. Importantly, our approach keeps the diffusion model frozen, ensuring computational efficiency while maintaining the strong generative quality of the pretrained model. 

As the storyline progresses, another challenge arises: the growing length of history increases computational and memory demands. Moreover, not all history pairs remain equally relevant, and earlier entries may introduce noise or dilute the focus of generation \cite{storygen}. To address this, we further introduce a salient history selection strategy that dynamically identifies the most informative history pair during inference. This improves generation quality by ensuring the model conditions only on the most relevant context.

To better evaluate how well the generated images reflect the narrative, we adopt TIFA~\cite{tifa}, a recently proposed metric that uses vision language models to verify whether specific concepts from the text appear in the image. Unlike CLIP-based similarity scores that primarily capture high-level semantic alignment, TIFA offers more targeted and interpretable assessments of text-image alignment by focusing on explicit concept grounding.

We conduct experiments on the StorySalon dataset and FlintStonesSV dataset, and compare ViSTA with state-of-the-art baselines. Results show that the story images generated by our model are not only consistent across frames, but also well-aligned with the narrative text. We summarize our contributions as follows:
\begin{enumerate}
    \item We propose a multi-modal history fusion model to extract and integrate relevant history fusion features from past text-image pairs, enabling effective use of history context for story generation.
    \item We utilize a lightweight history adapter that conditions the diffusion model on history fusion features without modifying its architecture or requiring full fine-tuning.
    \item We introduce a salient history selection strategy, which dynamically selects the most informative history during inference, improving the conditioning quality.
    \item We are the first to use TIFA to evaluate text-image alignment for visual storytelling, providing a more interpretable assessment of generated images.
\end{enumerate}

\section{Related Works}

\subsection{Visual Story Telling}
Visual storytelling involves generating a sequence of images based on narrative text prompts to create a consistent and coherent story.
Several recent works have explored an auto-regressive generation process, first introduced by AR-LDM \cite{arldm} where, in addition to the current text prompt, all history text-image pairs serve as additional conditions for generating the next image. Building upon this approach, Make-A-Story \cite{makeastory} proposes a visual memory-attention module to enhance consistency across frames. 

However, not all historical information should necessarily contribute equally to the generation process. To address this limitation, ACM-VSG \cite{acmvsg} introduces an adaptive encoder that evaluates the semantic relationship between the current caption and history text-image pairs. It further selects the most relevant history image as an additional guidance signal during sampling to improve global consistency. Causal-Story \cite{causalstory} takes a different approach by incorporating a local causal attention mechanism that quantifies the causal relationships between history and current text-image pairs, assigning appropriate weights to past information. StoryGen \cite{storygen} enhances temporal awareness by introducing different noise levels to history images, enabling the model to better distinguish temporal order.
To further improve the robustness of history features, StoryGPT-V \cite{shen2023storygptv} leverages LLMs for their strong reasoning capabilities, helping the model navigate ambiguous references. A character-aware latent diffusion model is then aligned with LLM outputs to generate character-consistent stories.

There are other methods \cite{zhu2023cogcartoon, talecrafter, storymaker, jeong2023zero, tao2024storyimager, zheng2024temporalstory, storynizor} that aim to improve story generation in different ways by introducing extra control, such as layout, stretch, identity embedding, \etc 
More recently, MM-Interleaved~\cite{tian2024mminterleaved} and SEED-Story~\cite{yang2024seedstory} employ a Multimodal Large Language Model (MLLM) to generate image features based on history text-image pairs and the current text, which are then used as conditions to guide the image generation for diffusion models. While this approach leverages the comprehension and generation ability of the MLLM to improve consistency, it significantly increases computational costs due to the need for fine-tuning both the MLLM and the diffusion model. 

Despite their effectiveness, auto-regressive methods have a fundamental limitation. Integrating history text-image pairs as additional conditioning signals requires architectural modifications to the U-Net within the diffusion model. These modifications often involve adding extra input channels and/or a different cross-attention mechanism to process the historical context effectively. As a result, these models require extensive training to adapt to the new architecture and learn how to leverage the additional historical information properly. This training process is computationally expensive, demanding large-scale datasets and significant fine-tuning efforts to achieve high-quality and consistent image generation. 

In contrast, we propose a multi-modal fusion model to extract and integrate history fusion features and propose a history adapter which can be seamlessly inserted into diffusion models to condition the history features. By keeping the diffusion model frozen while only training the fusion model and adapter, our method reduces the need for extensive training for the whole architecture. 

\begin{figure*}[]
    \centering
    \includegraphics[width=0.85\textwidth]{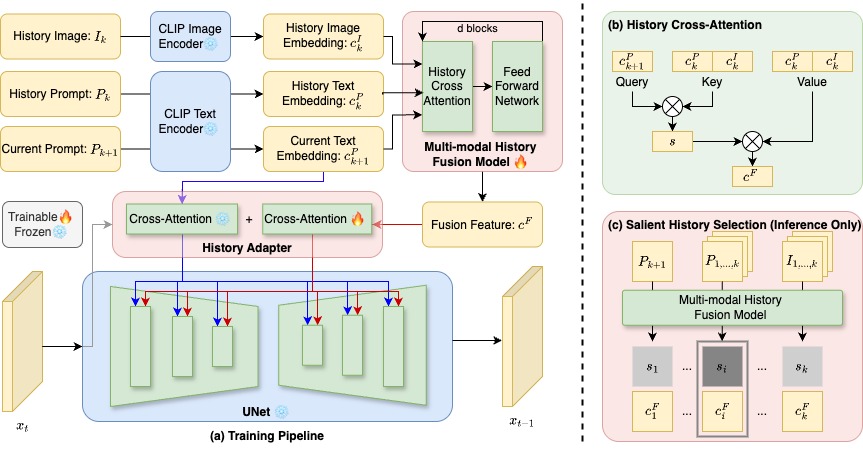}
    \caption{Overview of our proposed ViSTA model. (a) To generate the current image $I_{k+1}$, ViSTA takes text prompt $P_{k+1}$, history prompt $P_k$, and history image $I_k$ as input. We propose a multi-modal history fusion model to extract and integrate the history information and output a fusion feature $c^F$. We utilize a lightweight history adapter to condition on the fusion feature, avoiding modifying the diffusion models' architecture or full fine-tuning of the models; only the fusion model and adapter need to be trained. (b) Our proposed multi-modal history fusion model consists of $d$ blocks of history cross-attention and feedforward layers. In the history cross-attention, we use the current prompt embedding $c^P_{k+1}$ as the query, and the concatenation of history prompt $c^P_{k}$ and image embedding $c^I_{k}$ as the key and value. (c) Salient history selection strategy during inference. Instead of using all history prompts $P_{1, ..., k}$ and images $I_{1, ..., k}$ as references, we select the most salient history based on the attention score $s$. In this way, we make the model focus on the most informative history.}
    \label{fig:pipe}
\end{figure*}


\subsection{Subject Consistent Generation}
An alternative line of research focuses on subject-consistent generation, which aims to maintain visual consistency across images in a training-free, plug-and-play manner.

The Chosen One~\cite{thechosenone} and OneActor \cite{wang2024oneactor} propose cluster-based test-time tuning methods to optimize the character embeddings. 
ConsiStory~\cite{tewel_training-free_2024} proposes a subject-driven shared attention block and a correspondence-based feature injection mechanism to promote consistent appearance of subjects across frames.
StoryDiffusion~\cite{zhou2024storydiffusion} proposes Consistent Self-Attention that incorporates sampled reference tokens from the images generated within the same batch, to improve consistency. 
DreamStory \cite{he2024dreamstory} proposes a novel multi-subject consistent model that leverages the created character portraits of the subjects and their corresponding textual information. 
OnePromptOneStory~\cite{onepromptonestory} proposes concatenating all prompts as a single extended prompt and refining frame descriptions with Singular-Value Reweighting. They also propose an Identity-Preserving Cross-Attention to strengthen character consistency.

Subject-consistent generation methods typically adopt a more structured prompt format, such as ``character description + activity''. This format explicitly separates the character identity from the specific actions or scenes, making it easier for the model to preserve consistency. Repetition of the same character description across different prompts, the model reinforces key visual attributes throughout the sequence. 
However, while effective in preserving visual coherence, these methods come with several notable limitations.
(1) Lack of reference image support: they cannot generate specific predefined characters based on external visual inputs. Instead, they rely solely on textual descriptions, making it difficult to recreate a particular character with fine-grained control.
(2) To enforce consistency, the same character prompt must be used across all frames, which restricts the model’s ability to adapt dynamically to evolving narrative descriptions. This limitation is also problematic for complex storylines that involve multiple characters.

In contrast, we focus on the auto-regressive visual storytelling task, where each image is generated based on a natural language narrative that evolves over time. This setting allows for more expressive and flexible storytelling.

\section{Method}
We introduce the problem setup of the visual storytelling task in Section~\ref{sec:setup}. We then present the modules of our storytelling model ViSTA. We present ViSTA's multi-modal history fusion module to extract and integrate historical information in Section~\ref{sec:fusion}. We then illustrate how we utilize a lightweight history adapter for diffusion models to condition the fusion features in Section~\ref{sec:adapter}. Then, we present our salient history selection strategy in Section~\ref{sec:salient_hist_select}. Finally, we introduce how we utilize TIFA to evaluate the text-image alignment for visual storytelling task in Section~\ref{sec:tifa}. 
The overview of ViSTA's pipeline is shown in Figure~\ref{fig:pipe}. 

\subsection{Problem Setup}\label{sec:setup}

In this work, we focus on the visual storytelling task, where a sequence of images is generated based on narrative text captions. Given a sequence of text prompts that form a storyline, $P_1, P_2, \dots, P_K$, our goal is to generate a corresponding sequence of images, $I_1, I_2, \dots, I_K$. Following \cite{arldm, storygen}, we adopt an auto-regressive generation approach, \ie, at each image generation step, all previously generated text-image pairs, along with the current text caption, are used as input. 
We define the $k+1$-th generation step as:
$$I_{k+1} = \text{ViSTA} (P_{k+1}; P_{k}, I_{k}, \dots, P_1, I_1),$$

\noindent where $k \in \{1, ..., K\}$. Note that we focus on the story continuation setup, \ie, the first real image $I_1$ along with its caption $P_1$ is provided as the initial history to generate $I_2$. 

\subsection{Multi-modal History Fusion Model}\label{sec:fusion}
History text and images contain rich contextual information that can effectively guide the generation of the current image. For example, previous text and images may include overlapping characters, objects, or implicit references that contribute to continuity and coherence. Moreover, history images and their corresponding text naturally align, as the text describes the elements present in the images. Therefore, we propose a multi-modal fusion model to extract and integrate relevant information from history text-image pairs, ensuring a more consistent story generation. 
Specifically, we leverage the cross-attention mechanism to extract history fusion features.

Given the current text prompt $P_{k+1}$ and a history text-image pair $(P_{k}, I_{k})$, we first utilize pretrained CLIP image and text encoders to encode the text and image into their respective embeddings: $c^P_{k+1}$, $c^P_{k}$, and $c^I_{k}$. Note that projection layers are applied to align text and image embeddings to the same dimension.

Next, we employ a cross-attention mechanism where the current text embeddings serve as queries to attend to the concatenated history text and image embeddings. The resulting fusion feature $c^F$ is computed as:
\begin{equation}\label{eq:fusion}
c^F = \text{Attn}(Q^F,K^F,V^F)
= \text{Softmax} \left(\frac{Q^F(K^F)^T}{\sqrt{d}}\right) V^F,
\end{equation}
where the query, key, and value are obtained as follows:
\begin{equation}
    Q^F = c^P_{k+1} W_q^F, 
\end{equation}
\begin{equation}
    K^F = (c^P_{k} \| c^I_{k}) W_k^F, 
\end{equation}
\begin{equation}
    V^F = (c^P_{k} || c^I_{k}) W_v^F. 
\end{equation}
Here, the symbol $\|$ indicates concatenation along the sequence dimension. The trainable weight matrices $W_q^F$, $W_k^F$, $W_v^F$ are linear projection layers that transform the input embeddings into query, key, and value representations for the attention computation.

The output fusion feature $c^F$ is designed to incorporate relevant history features based on the current text prompt $P_{k+1}$. This process enables the fusion feature $c^F$ to selectively integrate the information from the history text-image pair that is most relevant to the current prompt. In this way, it can effectively capture contextual dependencies, ensuring that historical visual and textual features are aligned with the current text, thereby enhancing the consistency.

\subsection{History Adapter} \label{sec:adapter}
Lightweight adapters designed for specific tasks are used along with the base generative models in T2I generation to enhance the generation quality for the task. One such method is IP-Adapter~\cite{ip-adapter} that has demonstrated strong performance in generating consistent images conditioned on a reference image, using only lightweight trainable parameters. Inspired by this, we propose a history adapter that can be seamlessly integrated into a frozen diffusion model. The history adapter leverages the fusion feature $c^F$, which encodes relevant historical context, as an additional conditioning signal to guide the story image generation process.

To incorporate these features effectively, we introduce new cross-attention layers into the original U-Net~\cite{unet} architecture. These layers allow the image latents from the diffusion process to attend to the history fusion feature $c^F$. Specifically, we modify the standard cross-attention operation so that the U-Net’s intermediate latent representations serve as queries, while the history fusion feature $c^F$ provide the keys and values:
\begin{equation}
Z^c = \text{Attn}(Q^c, K^c, V^c) = \text{Softmax} \left( \frac{Q^c(K^c)^T}{\sqrt{d}} \right) V^c.
\end{equation}
Here, $Q^c$ is derived from the U-Net latent features, and $K^c$, $V^c$ are computed from the fusion feature $c^F$. This cross-attention produces a context vector $Z^c$, which encodes history-aware guidance.

To preserve the original behavior of the pretrained diffusion model while integrating historical context, we compute the final cross-attention output as a weighted combination of the original U-Net cross-attention output $Z$ and the history-guided output $Z^c$, where the scaling factor $\lambda$ controls the influence of the history adapter:
\begin{equation}
Z' = Z + \lambda \cdot Z^c,
\end{equation}

\textbf{Training Objective. }
During training, we optimize the parameters of the multi-modal history fusion model and the history adapter while keeping the pretrained diffusion model frozen. We reformulate the training dataset by extracting two consecutive pairs of text images to be a four-tuple ($P_{k}, I_{k}$, $P_{k+1}$, $I_{k+1}$). The model is trained end-to-end using the same reconstruction loss as diffusion models:
\begin{equation}
    \mathcal{L} = \mathbb{E}_{x_0, \epsilon, c^P, c^F, t} \left\| \epsilon - \epsilon_\theta(x_t, c^P, c^F, t)\right\|^2.
\end{equation}

\subsection{Salient History Selection}
\label{sec:salient_hist_select}
As the storyline progresses, the computational complexity and memory requirements increase. The model may also struggle to effectively focus on the most relevant historical context as the number of text-image pairs grows. In addition, earlier history pairs may become less relevant to the current generation, leading to potential inefficiencies in the conditioning process. Relying on all history text-image pairs as conditioning input can be computationally expensive and suboptimal. To address this, we propose a salient history selection strategy during inference. Instead of conditioning on all past text-image pairs, the salient history selection strategy computes the attention scores of all history pairs with respect to the current text prompt in ViSTA's fusion model, and selects the highest-scoring pair as the salient history for the history adapter.

When generating the $(k+1)$-th image, we have all previous $k$ text-image pairs $\{(P_1, I_1), ..., (P_k, I_k)\}$ as histories. 
Each pair of history and the current text prompt $P_{k+1}$ are fed into our history fusion model to get the fusion features $c^F_1, ...c^F_k$ and their corresponding attention scores $s_1, ..., s_k$. The attention score is calculated 
by:
\begin{equation}
    s_i = \text{Softmax} \left( \frac{Q^F_i (K^F_i)^T}{\sqrt{d}}\right).
\end{equation}
We select the most salient history fusion feature $c^F_{salient}$: 
\begin{equation}
    c^F_{salient} = \arg\max_i s_i.
\end{equation}

Instead of passing all the history to diffusion models, we only pass one fusion feature to mitigate the risk of information overload.
Our salient history selection strategy ensures that the diffusion model conditions on the most relevant history fusion feature.

\begin{figure}
    \centering
    \includegraphics[width=0.75\linewidth]{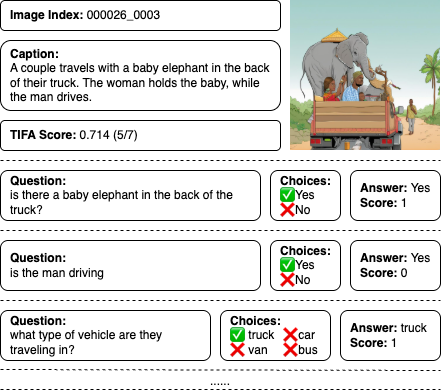}
    \caption{TIFA evaluation example. Given the caption, the following question-answer pairs are generated by a language model. Then a VQA model UnifiedQA evaluates the generated image based on the question-answer pairs. The final TIFA score for the generated image is the average on all questions.}
    \label{fig:tifa}
\end{figure}

\subsection{Text-Image Alignment Evaluation with TIFA}\label{sec:tifa}
In visual storytelling, narrative text prompts often describe specific objects, actions, and scenes that need to be accurately reflected in the generated images. Traditional evaluation metrics like CLIP similarity focus on overall semantic similarity but may overlook whether these specific elements are visually grounded. To better assess this alignment, we adapt TIFA~\cite{tifa}, a vision-language evaluation framework that measures how well an image supports the information described in its associated text.

We follow the question generation pipeline proposed in the original TIFA paper~\cite{tifa}, as shown in Figure~\ref{fig:tifa}. Specifically, a language model is used to automatically generate question-answer pairs from a given story caption, guided by in-context examples. These questions are designed to cover important content, including Yes / No questions (\eg, ``Is there a baby elephant in the back of the truck?'') and multiple choice questions (\eg, ``What type of vehicle are they traveling in? A. Truck, B. Car, C. Van, D. Bus''). These questions target specific entities and activities mentioned in the caption. 
Then, a Visual Question Answering (VQA) model is used to answer these questions based solely on the generated image. The TIFA score is calculated as the average accuracy across all questions, reflecting how well the image supports the textual content. 

In our work, we apply TIFA to evaluate whether the generated images faithfully represents the corresponding narrative prompt. To adapt TIFA to the storytelling setting, we treat each text and generated image pair independently and compute the score at the frame level. This provides a more targeted and interpretable evaluation of how well the generated visuals align with the story captions, especially when the story captions are long and rich in narrative structure.  
\begin{figure*}
    \centering    
    \begin{subfigure}{\textwidth}
    \centering
        \includegraphics[width=0.92\textwidth]{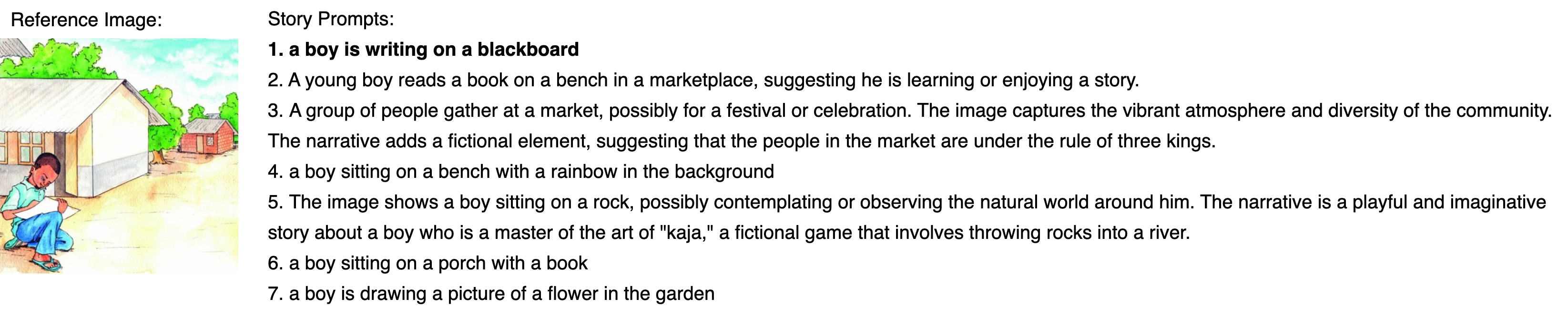}
    \end{subfigure}
    \begin{subfigure}{\textwidth}
        \raisebox{1.2cm}{\rotatebox[origin=c]{90}{Ground Truth}}
        \hfill
        \includegraphics[width=0.95\textwidth]{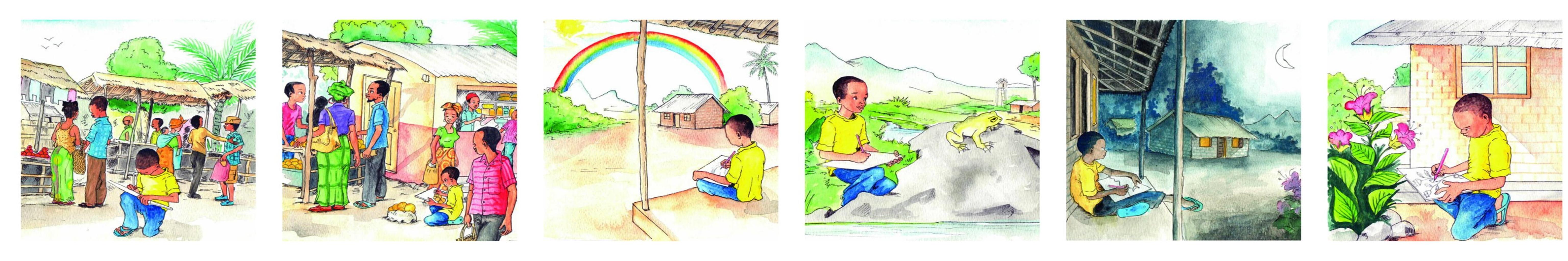}
    \end{subfigure}
    \begin{subfigure}{\textwidth}
        \raisebox{1.2cm}{\rotatebox[origin=c]{90}{SDXL-Prompt}}
        \hfill
        \includegraphics[width=0.95\textwidth]{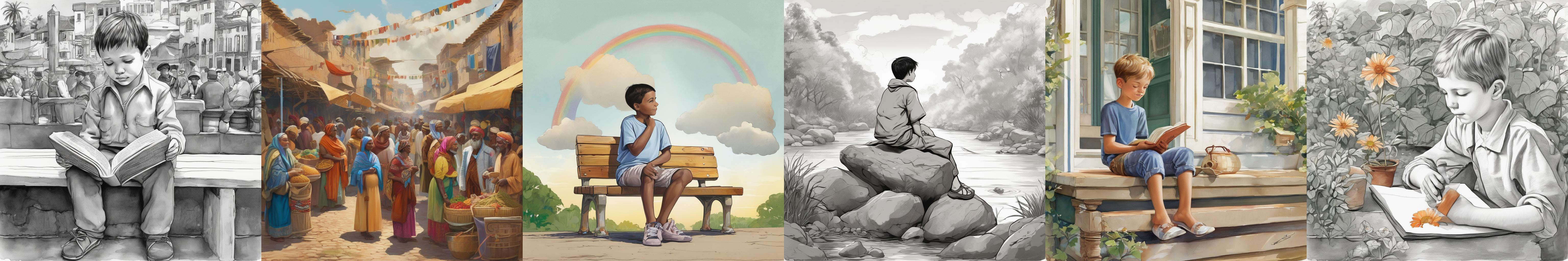}
    \end{subfigure}
    \begin{subfigure}{\textwidth}
        \raisebox{1.2cm}{\rotatebox[origin=c]{90}{IP-Adapter}}
        \hfill
        \includegraphics[width=0.95\textwidth]{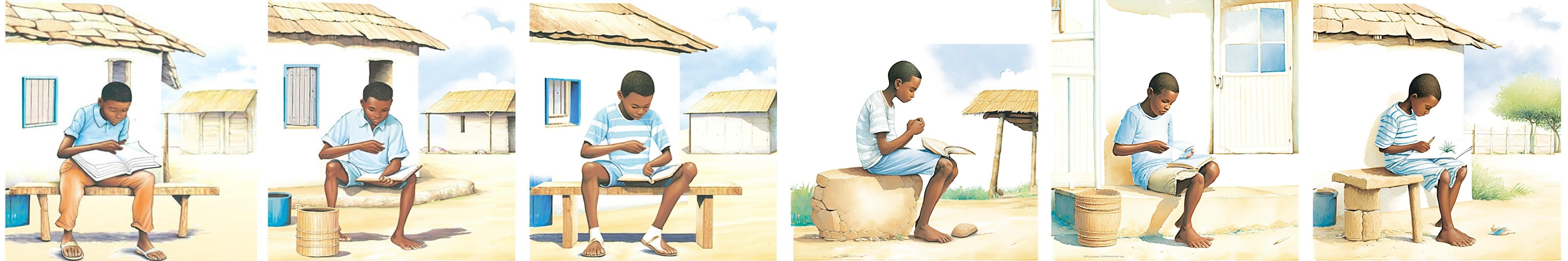}
    \end{subfigure}
    \begin{subfigure}{\textwidth}
        \raisebox{1.2cm}{\rotatebox[origin=c]{90}{StoryGen}}
        \hfill
        \includegraphics[width=0.95\textwidth]{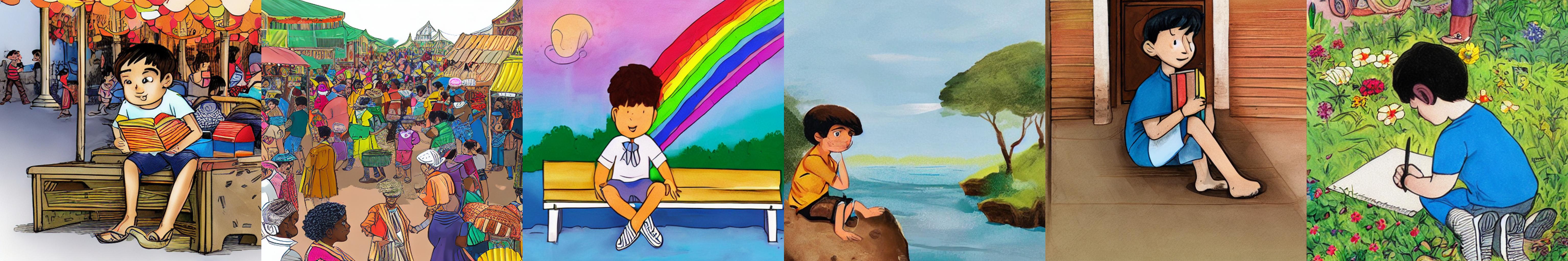}
    \end{subfigure}
    \begin{subfigure}{\textwidth}
        \raisebox{1.2cm}{\rotatebox[origin=c]{90}{Ours}}
        \hfill
        \includegraphics[width=0.95\textwidth]{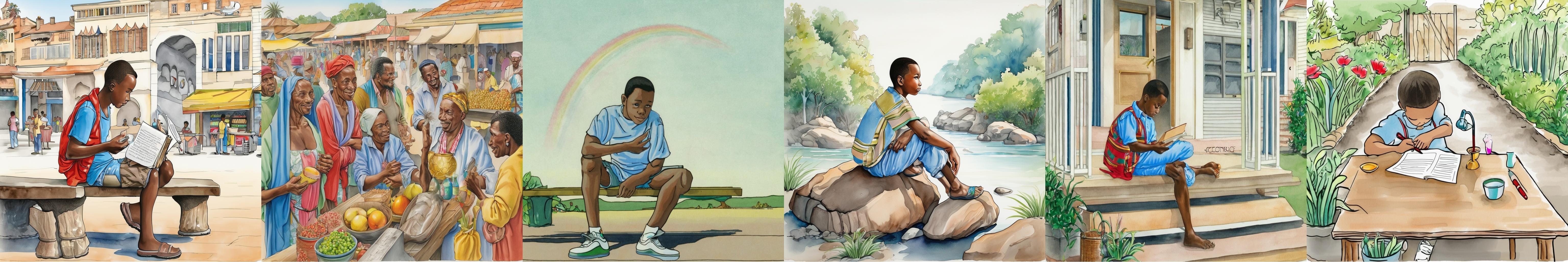}
    \end{subfigure}
    \caption{\textbf{Qualitative results: StorySalon.} This figure presents a comparison of storytelling between ViSTA, baseline methods, and state-of-the-art on a sample StorySalon story. We include ground truth images from the dataset as an evaluation for visual coherence and accuracy. The same reference image is used across all methods. The first story prompt corresponds to the reference image, while story prompts 2 to 7 are used to generate the subsequent frames. While SDXL-Prompt show high-quality and well-aligned images, they fail in generating consistent character across all frames. Although IP-Adapter shows consistent character, the generated images do not align with the prompt. Compare with the state-of-the-art StoryGen, our ViSTA shows better consistency on both characters and style.}
    \label{fig:compare}
    \vspace{-0.4cm}
\end{figure*}

\section{Experiments}
We evaluate our proposed method on two different dataset with different settings, StorySalon~\cite{storygen} which does not contain recurring characters, while FlintStonesSV~\cite{flintstonessv} has seven recurring characters. 
For both datasets, the first text-image pair is given as an initial reference. Afterwards, the generated frames serve as history for further generations. 
Experimental setup, including dataset descriptions, evaluation metrics, baselines, implementation details, as well as the code can be found in the supplementary material. 

\begin{table*}[]
    \centering
    \begin{tabular}{l|c|c|c|c|c}
    \toprule
    & \multicolumn{2}{|c|}{Text-Image Alignment} & \multicolumn{2}{|c|}{Image-Image Alignment} & \\
    \cline{2-3} \cline{4-5}
    \multicolumn{1}{l|}{Model} & \multicolumn{1}{|c|}{CLIP-T $\uparrow$} & \multicolumn{1}{|c|}{TIFA $\uparrow$} & \multicolumn{1}{|c|}{CLIP-I $\uparrow$} & \multicolumn{1}{|c|}{FID $\downarrow$} & \multicolumn{1}{|c}{Inference Time (s)} \\
    \midrule
    SDXL-Prompt & \textbf{\textcolor{blue}{0.319}} & \textbf{\textcolor{blue}{0.782}} & \textcolor{red}{0.697} & \textcolor{red}{63.66} & 5.83 \\
    IP-Adapter  & \textcolor{red}{0.255} & \textcolor{red}{0.521} & \textbf{\textcolor{blue}{0.725}} & \textbf{\textcolor{blue}{42.66}} & 6.41 \\
    \hdashline
    StoryGen~\cite{storygen}    & 0.313 & 0.750 & 0.710 & 47.28 & 8.23 \\
    ViSTA (ours)        & \underline{0.316} & \underline{0.765} & \underline{0.715} & \underline{46.45} & 6.33\\
    \bottomrule    
    \end{tabular}
\caption{\textbf{Quantitative results.} We evaluate text-image alignment and image-image consistency on the E-Book partition of the StorySalon test set. The SDXL-Prompt baseline, which generates images using only the current text prompt with an animation prefix, achieves the highest text-image alignment scores (CLIP-T and TIFA). However, it suffers from the lowest image-image similarity (CLIP-I) and fidelity (FID) due to the lack of visual conditioning--\ie, SDXL model only takes the current prompt as input.
In contrast, IP-Adapter, which conditions image generation solely on history images along with the current prompt, significantly improves image-image similarity and FID, as it directly references visual appearance from previous frames. However, it performs worse on text-image alignment because it lacks access to the full narrative prompt and relies only on visual cues without detailed textual guidance.
These results highlight the importance of conditioning on both history text and image features to achieve a balance between visual consistency and narrative relevance. Our proposed ViSTA model, which leverages both modalities, outperforms the state-of-the-art StoryGen~\cite{storygen} on all evaluation metrics.
Blue bold numbers indicate the best scores, red numbers indicate the worst, and underlined numbers represent state-of-the-art result for storytelling.}
\label{tab:main}
\end{table*}

\subsection{Results}

Table~\ref{tab:main} presents the quantitative results on the E-book partition of the StorySalon test dataset. As shown, SDXL-Prompt achieves the highest scores on CLIP-T and TIFA, demonstrating strong alignment with the input text prompt. However, without access to the history image or text, SDXL-Prompt performs poorly on CLIP-I and FID, indicating low visual consistency across frames. In contrast, IP-Adapter, which is conditioned solely on history images, achieves significantly better image-image similarity and fidelity but at the cost of degraded text-image alignment, due to the absence of detailed textual guidance of the history. The state-of-the-art method, StoryGen, provides a more balanced performance, but it still struggles to fully leverage both text and image history.

Our proposed model, ViSTA, which conditions history text and image, outperforms StoryGen, a state-of-the-art model, across all metrics, demonstrating the effectiveness in preserving both text alignment and image consistency. Moreover, ViSTA employs a lightweight adapter without modifying the diffusion model architecture or requiring full fine-tuning, further highlighting its efficiency and practical applicability in generating high-quality story images.

We also compared the inference time (in seconds) for each method under the same hardware setting (one L40S GPU). Note that StoryGen is based on SD v1.5, which uses a smaller backbone than SDXL but incorporates an auto-regressive generation module that has a significant 7.3x increase on inference time. 
In contrast, our method is built on SDXL and introduces only a lightweight history adapter, resulting in a minimal increase in runtime--just 1.1x (0.5 seconds) more than vanilla SDXL. Compared to  IP-Adapter, ViSTA achieves comparable efficiency while delivering improved character consistency and narrative alignment.


Figure~\ref{fig:compare} shows the qualitative results of ViSTA, baselines, and state-of-the-art methods. 
The same reference image is initially used across IP-Adapter, StoryGen, and ViSTA. The first story prompt is the corresponding caption for the reference image and is used for both StoryGen and ViSTA. 
We also include the ground truth images from the StorySalon dataset, which serve as the reference for evaluating the coherence and accuracy of the generated story images. These real images illustrate how the same character and style evolve naturally throughout the story, providing a target for visual and narrative consistency. We observe that although SDXL-Prompt generates high-quality and well-aligned images, it struggles to maintain character consistency across frames. In contrast, IP-Adapter ensures consistent character appearances but lacks strong text-image alignment, often failing to accurately reflect the given prompts. Compared to the state-of-the-art StoryGen, ViSTA achieves superior consistency in both character identity and visual style, effectively improving text alignment and consistency throughout the generated image sequence.

Table~\ref{tab:flintstones} presents quantitative results on the FlintstonesSV test dataset. 
Our observations on FlintstonesSV align closely with previous findings on SDXL-Prompt and IP-Adapter, demonstrating ViSTA's effectiveness in preserving both text alignment and image consistency. 
Compared to the other state-of-the-art methods, StoryGPT-V~\cite{shen2023storygptv} 
, ViSTA achieves superior results, indicating notably improved quality in terms of image realism and distribution similarity.
More results are presented in the appendix. 

\begin{table}[]
\vspace{-0.5cm}
    \setlength{\tabcolsep}{2pt}
    \centering
    \small
    \begin{tabular}{l|c|c|c|c}
    \toprule
    & \multicolumn{2}{c|}{Text-Image Align.} & \multicolumn{2}{|c}{Image-Image Align.} \\
    \cline{2-3} \cline{4-5}
    \multicolumn{1}{l}{Model} & \multicolumn{1}{c}{CLIP-T $\uparrow$} & \multicolumn{1}{c|}{TIFA $\uparrow$} & \multicolumn{1}{c}{CLIP-I $\uparrow$} & \multicolumn{1}{c}{FID $\downarrow$} \\
    \midrule
    SDXL-Prompt & \textbf{\textcolor{blue}{0.276}} & \textbf{\textcolor{blue}{0.685}} & \textcolor{red}{0.724} & 10.00 \\
    IP-Adapter  & \textcolor{red}{0.253} & \textcolor{red}{0.511} & 0.833 & \textbf{\textcolor{blue}{5.87}} \\
    \hdashline
    StoryGPT-V~\cite{shen2023storygptv} & -- & -- & -- & \textcolor{red}{21.13} \\
    ViSTA (ours)        & \underline{0.266} & \underline{0.558} & \textbf{\textcolor{blue}{\underline{0.843}}} & \underline{8.00} \\
    \bottomrule    
    \end{tabular}
\caption{Quantitative results: FlintStonesSV. StoryGPT-V numbers are reported in \cite{shen2023storygptv}. 
Results are consistent with prior SDXL-Prompt and IP-Adapter observations. ViSTA outperforms SOTA StoryGPT-V~\cite{shen2023storygptv} on FID on the recurring characters setting, indicating improved image quality.
}
\vspace{-0.3cm}
\label{tab:flintstones}
\end{table}



\subsection{Human Evaluation} 
We conducted a pilot study to rank the story generation methods on three main criteria: (1) \textit{Textual Alignment:} how well does each frame follow its corresponding prompt (2) \textit{Image Alignment:} how well does each frame follow the overall style of the reference image, and (3) \textit{Consistency:} how well is the frame-to-frame consistency (same characters/objects). 
We asked five non-author graduate students to rank five examples per person. Table~\ref{tab:human_evaluation} shows the elo ranking. The initial elo score was 128, and the tuning factor is 4. It is also worth noting that our method is the top 1 ranked method 69.23\%, 57.69\%, and 50.00\% of the time when compared with the other 3 methods in Textual Alignment, Image Alignment, and Consistency, respectively. More detail can be found in the supplementary material.

\begin{table}[]
\vspace{-0.5cm}
\setlength{\tabcolsep}{2pt}
\centering
\small
\begin{tabular}{l|c|c|c}
\toprule
Model & Textual Align. & Image Align. & Consistency \\
\midrule
SDXL-Prompt    & 174.78 & 128.51 & 117.77 \\
IP-Adapter     & 8.62   & 81.21  & 178.05 \\
StoryGen       & 107.71 & 85.00  & 26.03  \\
ViSTA (ours)   & \textbf{220.88} & \textbf{217.28} & \textbf{190.15} \\
\bottomrule
\end{tabular}
\caption{\textbf{User Study.} Elo ranking for all methods across Textual Alignment, Image Alignment, and Consistency metrics. 
}
\label{tab:human_evaluation}
\vspace{-0.3cm}
\end{table}

\section{Conclusion}
In this paper, we present ViSTA, a novel approach for visual storytelling that enhances character consistency and text-image alignment in text-to-image diffusion models. To effectively utilize history context, we introduce a multi-modal history fusion model that extracts and integrates relevant features from past text-image pairs, enabling coherent story generation. Additionally, we propose a lightweight history adapter that conditions the diffusion model on history fusion features without modifying its architecture or requiring full fine-tuning, ensuring efficiency and adaptability.
To further improve the generation process, we introduce a salient history selection strategy that dynamically identifies the most informative history text-image pair at each step, enhancing the quality of conditioning. Furthermore, we are the first to employ TIFA as a metric for evaluating text-image alignment in visual storytelling, providing a more targeted and interpretable assessment of generated images.
Extensive experiments on two datasets demonstrate that ViSTA outperforms state-of-the-art methods, achieving superior character consistency and text-image alignment.

{
    \small
    \bibliographystyle{ieeenat_fullname}
    \bibliography{main}
}

\input{appendix}

\end{document}

%% file: preamble.tex
\usepackage{subcaption}
\usepackage{arydshln}
\usepackage{xspace} 

\usepackage[most]{tcolorbox}

%% file: appendix.tex
\clearpage
\appendix

\section{Experimental Setup}

\subsection{Datasets}
We conduct our experiments on two dataset, StorySalon~\cite{storygen} and FlintStonesSV~\cite{flintstonessv}. 
The StorySalon dataset~\cite{storygen} is the largest and most recent benchmark for visual storytelling. It contains 159,778 animation-style images across 446 character categories. Designed to support long-range story generation, each story consists of an average of 14 frames, and each corresponding text prompt contains an average of 106 words. The dataset is constructed from E-books and YouTube videos retrieved using keyword-based queries.
To generate the narrative-aligned text descriptions for each image, TextBind~\cite{textbind} is applied to produce captions conditioned on both the image and the surrounding narrative text.
Due to a portion of the YouTube videos becoming unavailable, we conduct all experiments on the E-books partition, which remains complete and high-quality. It includes 8,635 training stories with 118,892 frames, and 451 test stories with 6,026 frames.

The FlintstonesSV dataset \cite{flintstonessv} serves as an additional benchmarking dataset in our experiments. It comprises 25,184 densely annotated, animation-style video clips sourced from the animated series ``The Flintstones.'' FlintstonesSV selects a single representative frame from each clip and groups frames from consecutive clips into coherent stories of length five. The dataset consists of seven recurring characters, offering a robust setting for evaluating character and scene consistency in visual storytelling tasks. FlintstonesSV is split into 20,132 training, 2,071 validation, and 2,309 test stories. 

\subsection{Evaluation Metrics}
To evaluate text-image alignment, we utilize CLIP text-image similarity (CLIP-T~\cite{clip}) and TIFA~\cite{tifa} score. Following the literature~\cite{acmvsg, arldm, storygen}, we use CLIP-T to evaluate high-level semantic text-image similarity. We also propose to use TIFA to assess the how well the generated images align with the narrative story text. Following~\cite{tifa}, we use GPT-3.5~\cite{gpt} to generate question-answer pairs and use UnifiedQA~\cite{unifiedqa} to evaluate generated images.
\footnote{\url{https://github.com/Yushi-Hu/tifa}}

To evaluate the quality of the generated images with respect to the ground-truth, we use Frechet Inception Distance (FID~\cite{fid}) score and CLIP image-image similarity (CLIP-I~\cite{clip}). FID is used to evaluate the quality of generated images by comparing the distributions of generated images and ground-truth images, quantifying the faithfulness and diverseness of generated images. CLIP-I measures the similarity between the generated and ground-truth images.

\subsection{Baselines}
For StorySalon dataset, we compare our method with the following: (1) SDXL-Prompt: 
\footnote{\url{https://huggingface.co/stabilityai/stable-diffusion-xl-base-1.0}} 
We use the pretrained Stable Diffusion XL model \cite{sdxl} given the current text as prompt with a ``A cartoon style image'' prefix, without using any history image or text.
(2) IP-Adapter \cite{ip-adapter}: We use the IP-Adapter for SDXL 
\footnote{\url{https://huggingface.co/h94/IP-Adapter/tree/main/sdxl_models}} 
to generate the image given the current text as its prompt, and all the previous generated history images as its reference images. 
(3) State-of-the-art StoryGen \cite{storygen}: We run the inference using the public checkpoint of the StoryGen model.
\footnote{\url{https://huggingface.co/haoningwu/StoryGen/tree/main/checkpoint_StorySalon}} 
For FlintStonesSV datset, we compare our method with SDXL-Prompt, IP-Adapter and StoryGPT-V~\cite{shen2023storygptv}.  

While a wide range of methods exist for visual storytelling, many rely on assumptions or constraints that are incompatible with our broader setting.

Specifically, we compare against StoryGen on the StorySalon dataset and StoryGPT-V on the FlintstonesSV dataset, which are, to the best of our knowledge, the only methods that operate under our broader setting.

Other methods were not included for the following reasons:
(1) StoryGPT-V leverages predefined recurring characters specific to the FlintstonesSV dataset, a feature not available in the StorySalon dataset. This makes it inapplicable for evaluation on StorySalon.
(2) We also excluded certain training-free methods (e.g., StoryDiffusion) because they are designed for short, ``character + activity'' format prompts and rely on repeated character prompt for consistency. We did experiments on StorySalon dataset with StoryDiffusion and OnePromptOneStory methods in our setting but yielded poor results. We believe that such comparisons would not be fair or meaningful.

Therefore, to our best knowledge, we chose the baselines that do not make such assumptions and are therefore applicable to the datasets we compare against.

\subsection{Implementation Details}
Our multi-modal history fusion model consists of $d=4$ blocks, each block contains a cross-attention layer and a FeedForward Network. The hidden dimension of the fusion model is 1024. We use OpenCLIP-ViT-H~\cite{openclip} as the image and text encoders. Our history adapter is built on the Stable Diffusion XL. We train the model end-to-end for 80,000 steps with a batch size of 4. We use AdamW optimizer~\cite{adamw} with a learning rate of 1e-4. During inference, we adopt a DDIM sampler~\cite{ddim} with 50 inference steps, and the guidance scale is 5.
All experiments are conducted on 4 NVIDIA L40S GPUs. 
We set $\lambda = 0.5$ to balance image quality and consistency. The larger $\lambda$ improves consistency but reduces quality, while the smaller $\lambda$ improves quality at the cost of consistency. The value $0.5$ offers the best trade-off in our experiments.

\section{Additional Results} 
\subsection{StorySalon Dataset}
\begin{figure*}
    \centering
    \begin{subfigure}{\textwidth}
    \centering
        \includegraphics[height=3.3cm]{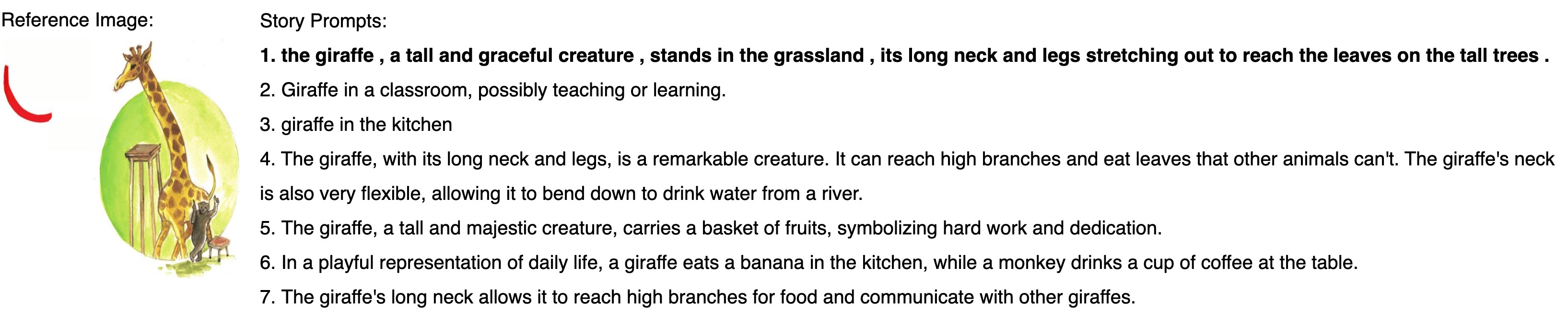}
    \end{subfigure}
    \begin{subfigure}{\textwidth}
        \raisebox{1.2cm}{\rotatebox[origin=c]{90}{Ground Truth}}
        \hfill
        \includegraphics[width=0.97\textwidth]{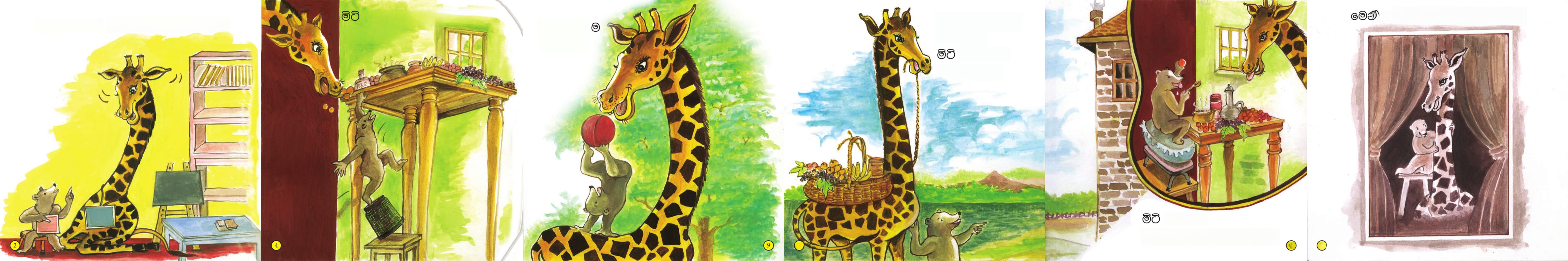}
    \end{subfigure}

    \begin{subfigure}{\textwidth}
        \raisebox{1.2cm}{\rotatebox[origin=c]{90}{SDXL-Prompt}}
        \hfill
        \includegraphics[width=0.97\textwidth]{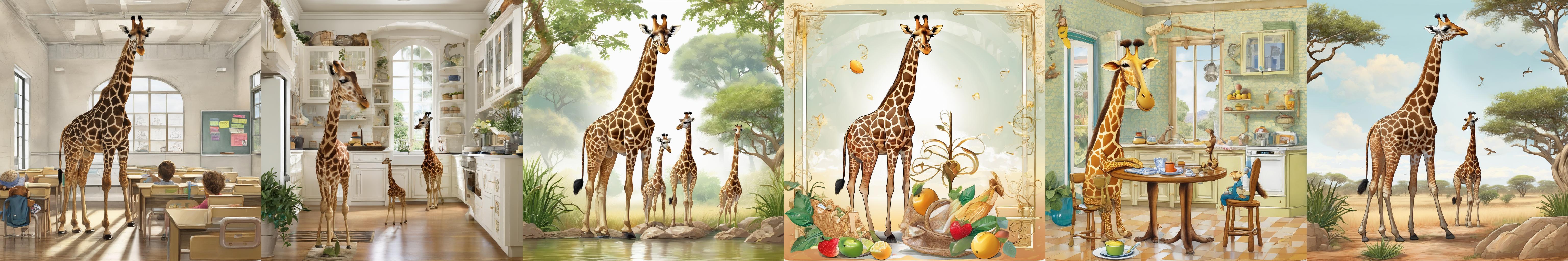}
    \end{subfigure}

    \begin{subfigure}{\textwidth}
        \raisebox{1.2cm}{\rotatebox[origin=c]{90}{IP-Adapter}}
        \hfill
        \includegraphics[width=0.97\textwidth]{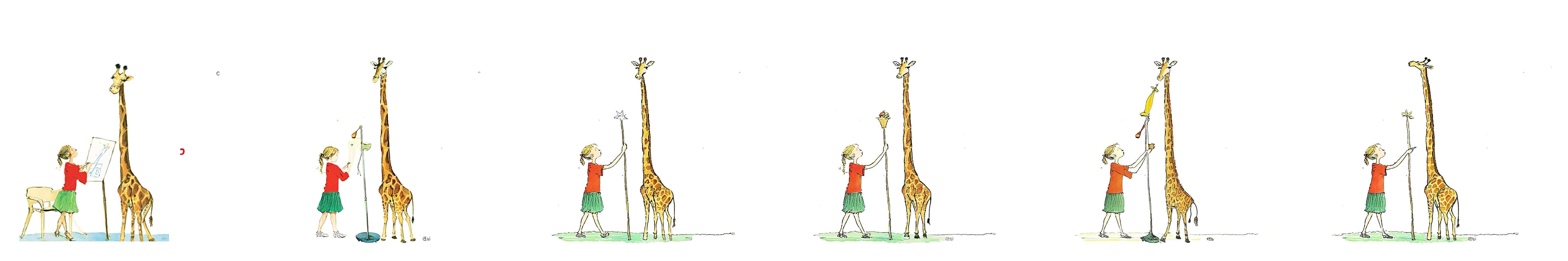}
    \end{subfigure}

    \begin{subfigure}{\textwidth}
        \raisebox{1.2cm}{\rotatebox[origin=c]{90}{StoryGen}}
        \hfill
        \includegraphics[width=0.97\textwidth]{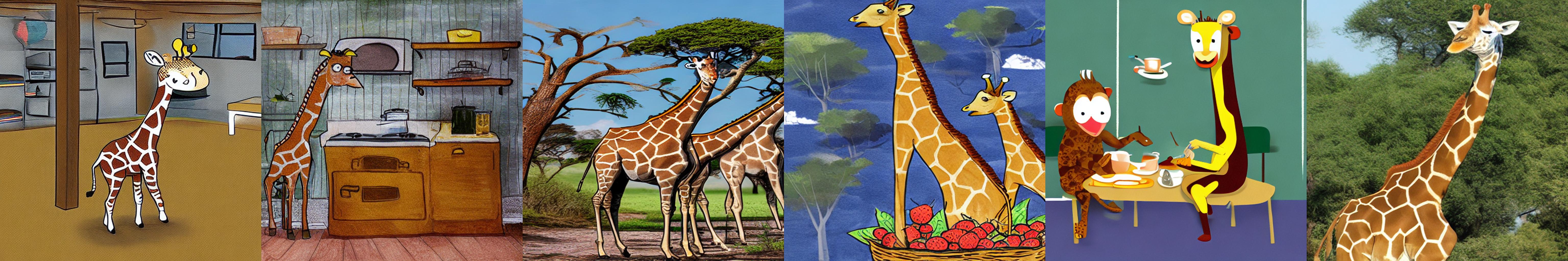}
    \end{subfigure}

    \begin{subfigure}{\textwidth}
        \raisebox{1.2cm}{\rotatebox[origin=c]{90}{Ours}}
        \hfill
        \includegraphics[width=0.97\textwidth]{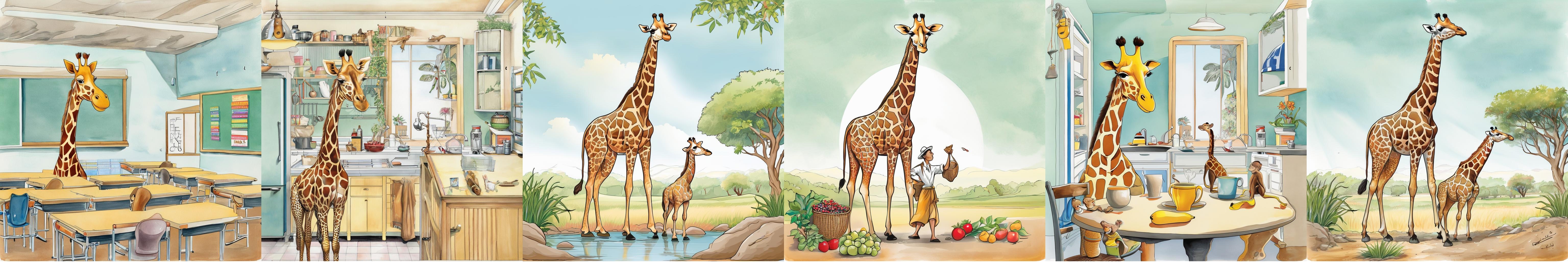}
    \end{subfigure}

    \caption{\textbf{Additional Qualitative results.} This figure presents a comparison of storytelling between ViSTA, baseline methods, and state-of-the-art on a sample StorySalon story. While SDXL-Prompt show high-quality and well-aligned images, they fail in generating consistent character acorss all frames. Although IP-Adapter shows consistent character, the generated images do not align with the prompt. Compare with the state-of-the-art StoryGen, our ViSTA shows better consistency on both characters and style.}
    \label{fig:x1}
\end{figure*}

\begin{figure*}
    \centering    
    \begin{subfigure}{\textwidth}
    \centering
        \includegraphics[height=3.1cm]{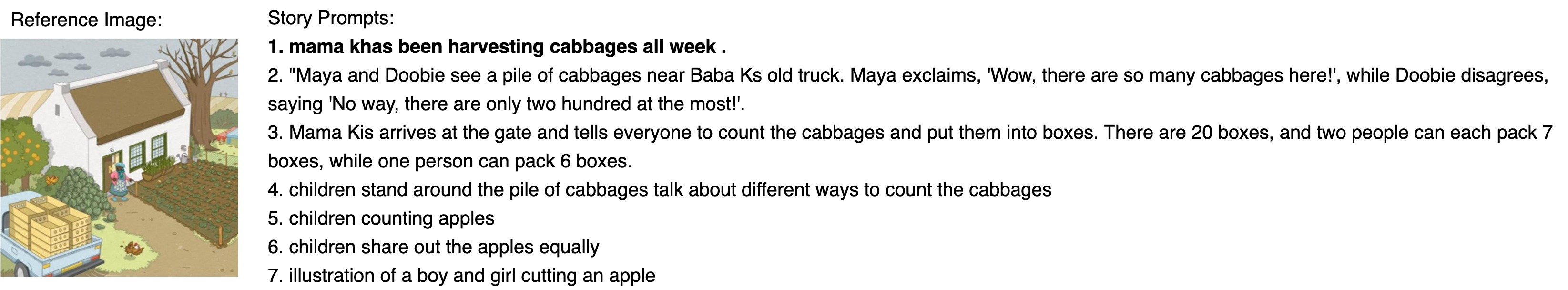}
    \end{subfigure}
    \begin{subfigure}{\textwidth}
        \raisebox{1.2cm}{\rotatebox[origin=c]{90}{Ground Truth}}
        \hfill
        \includegraphics[width=0.97\textwidth]{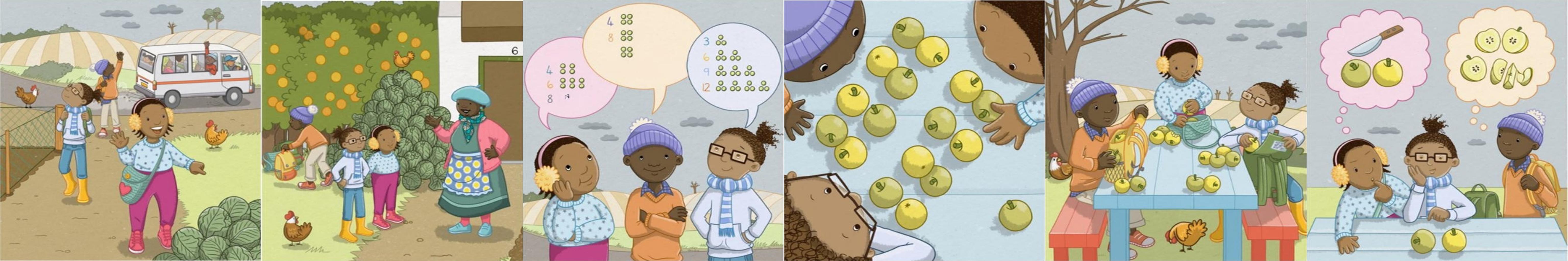}
    \end{subfigure}

    \begin{subfigure}{\textwidth}
        \raisebox{1.2cm}{\rotatebox[origin=c]{90}{SDXL-Prompt}}
        \hfill
        \includegraphics[width=0.97\textwidth]{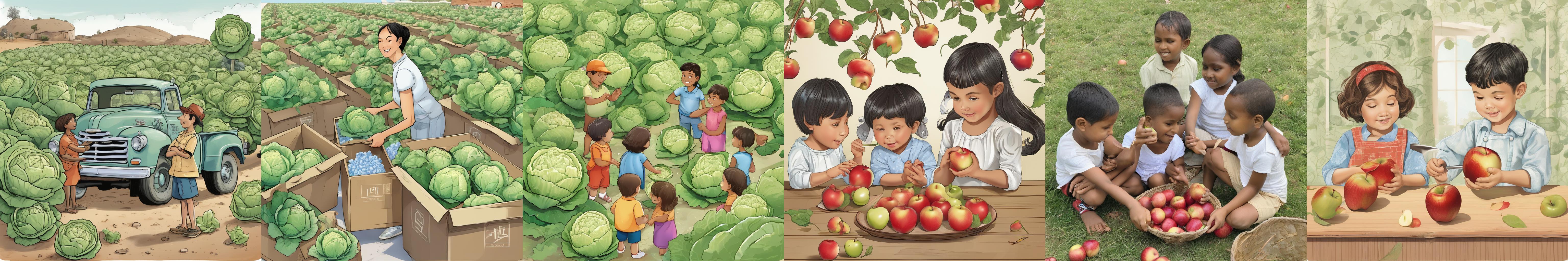}
    \end{subfigure}

    \begin{subfigure}{\textwidth}
        \raisebox{1.2cm}{\rotatebox[origin=c]{90}{IP-Adapter}}
        \hfill
        \includegraphics[width=0.97\textwidth]{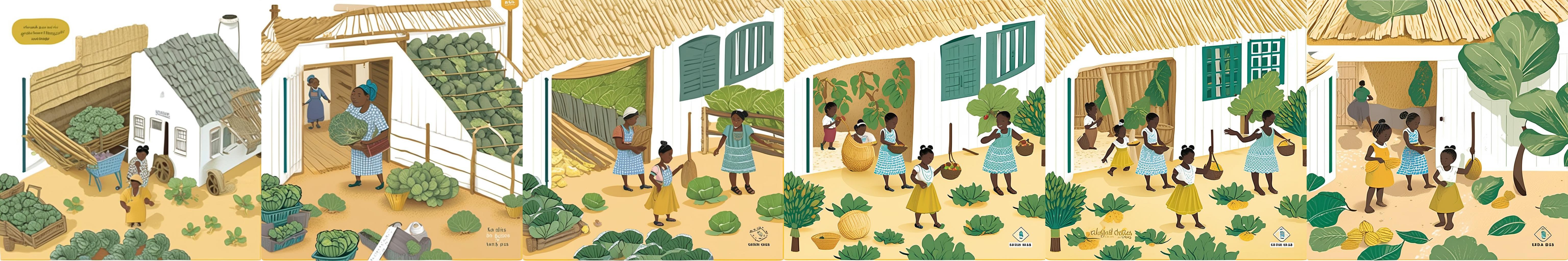}
    \end{subfigure}

    \begin{subfigure}{\textwidth}
        \raisebox{1.2cm}{\rotatebox[origin=c]{90}{StoryGen}}
        \hfill
        \includegraphics[width=0.97\textwidth]{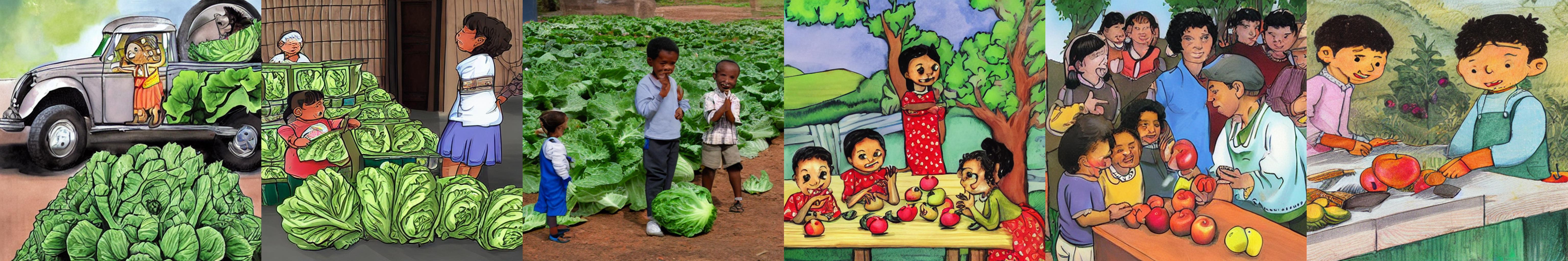}
    \end{subfigure}

    \begin{subfigure}{\textwidth}
        \raisebox{1.2cm}{\rotatebox[origin=c]{90}{Ours}}
        \hfill
        \includegraphics[width=0.97\textwidth]{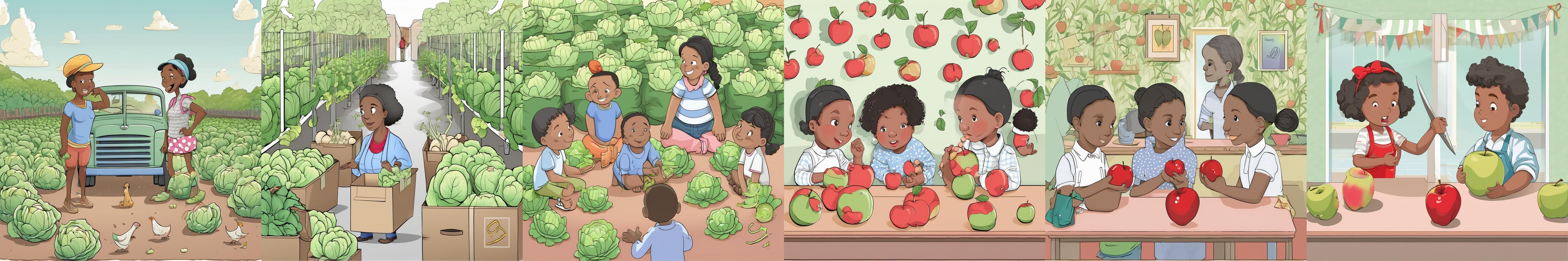}
    \end{subfigure}

    \caption{\textbf{Additional Qualitative results.} This figure presents a comparison of storytelling between ViSTA, baseline methods, and state-of-the-art on a sample StorySalon story. While SDXL-Prompt show high-quality and well-aligned images, they fail in generating consistent character acorss all frames. Although IP-Adapter shows consistent character, the generated images do not align with the prompt. Compare with the state-of-the-art StoryGen, our ViSTA shows better consistency on both characters and style.}
    \label{fig:x2}
\end{figure*}

\begin{figure*}[h]
    \centering
    \begin{subfigure}{\textwidth}
    \centering
        \includegraphics[width=0.92\linewidth]{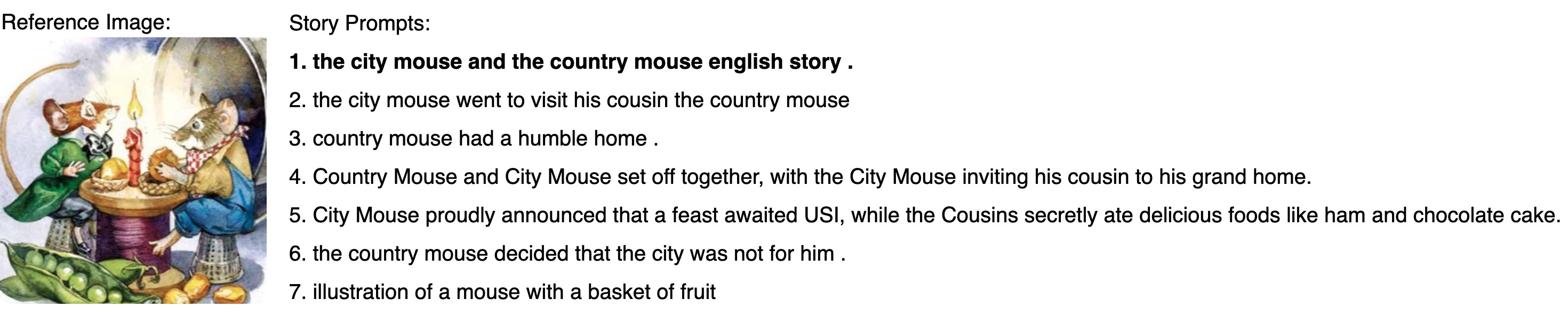}
    \end{subfigure}
    \begin{subfigure}{\textwidth}
        \raisebox{1.2cm}{\rotatebox[origin=c]{90}{Ground Truth}}
        \hfill
        \includegraphics[width=0.95\linewidth]{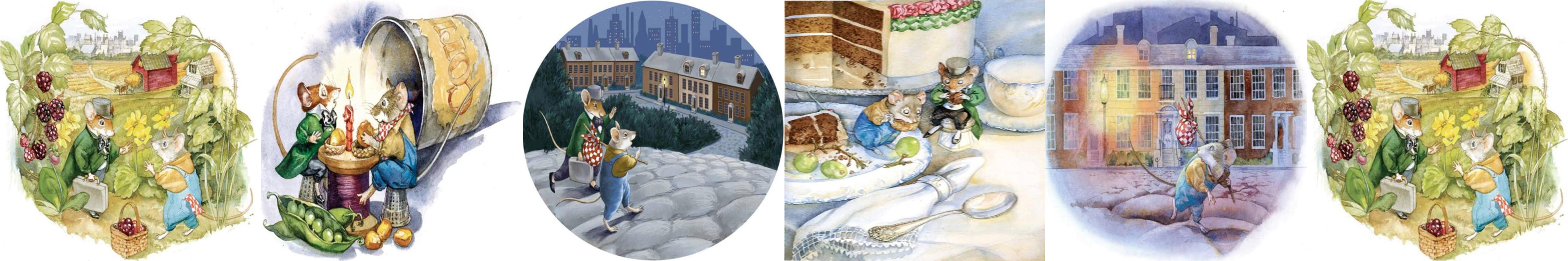}
    \end{subfigure}
    \begin{subfigure}{\textwidth}
        \raisebox{1.2cm}{\rotatebox[origin=c]{90}{SDXL-Prompt}}
        \hfill
        \includegraphics[width=0.95\linewidth]{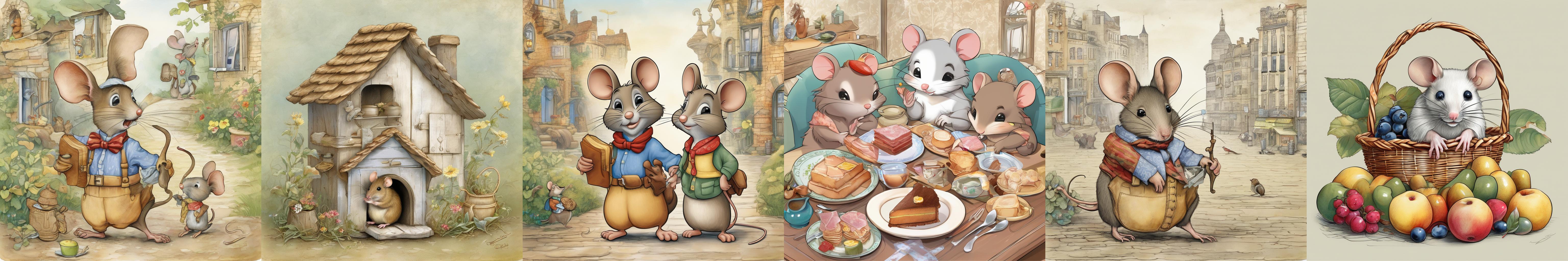}
    \end{subfigure}
    \begin{subfigure}{\textwidth}
        \raisebox{1.2cm}{\rotatebox[origin=c]{90}{IP-Adapter}}
        \hfill
        \includegraphics[width=0.95\linewidth]{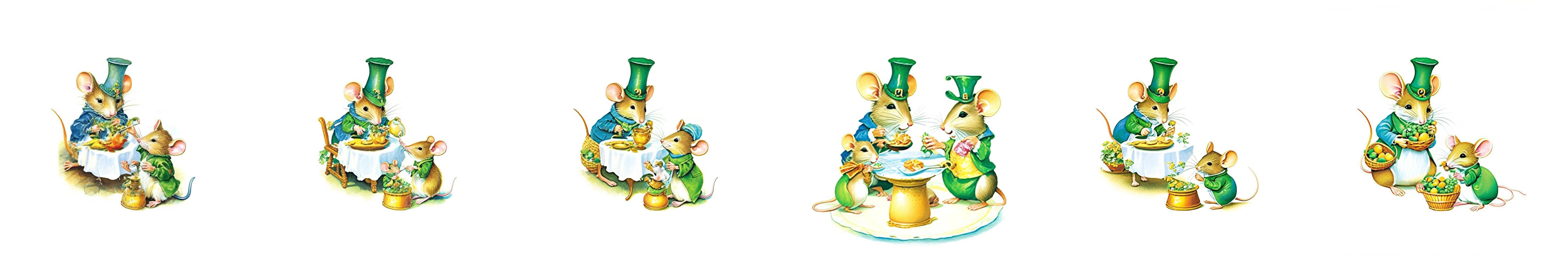}
    \end{subfigure}
    \begin{subfigure}{\textwidth}
        \raisebox{1.2cm}{\rotatebox[origin=c]{90}{StoryGen}}
        \hfill
        \includegraphics[width=0.95\linewidth]{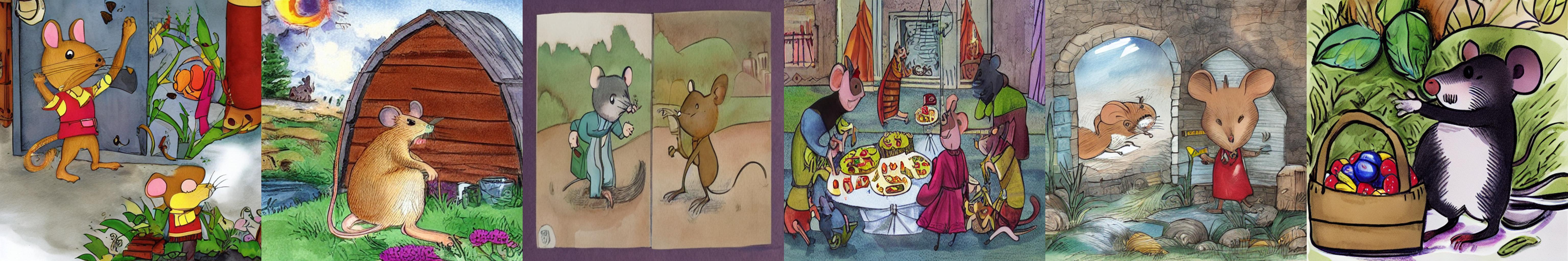}
    \end{subfigure}
    \begin{subfigure}{\textwidth}
        \raisebox{1.2cm}{\rotatebox[origin=c]{90}{Ours}}
        \hfill
        \includegraphics[width=0.95\linewidth]{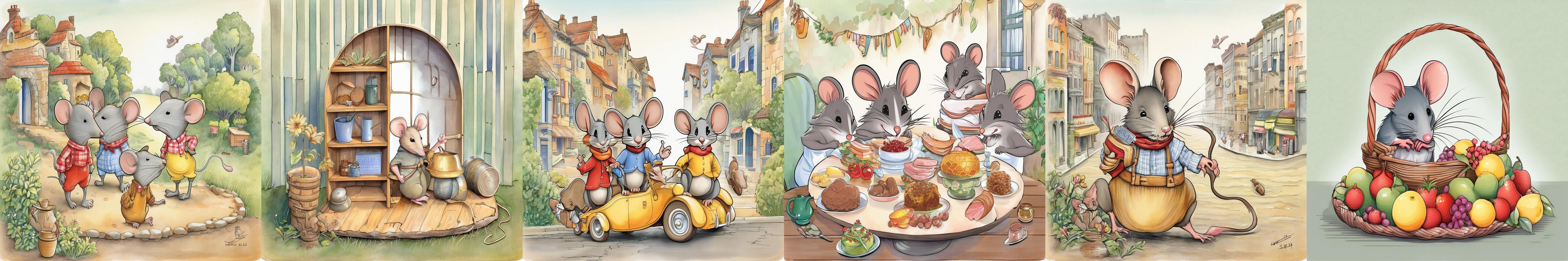}
    \end{subfigure}
    \caption{\textbf{Additional Qualitative results.} This figure presents a comparison of storytelling between ViSTA, baseline methods, and state-of-the-art on a sample StorySalon story. While SDXL-Prompt show high-quality and well-aligned images, they fail in generating consistent character acorss all frames. Although IP-Adapter shows consistent character, the generated images do not align with the prompt. Compare with the state-of-the-art StoryGen, our ViSTA shows better consistency on both characters and style.}
    \label{fig:x3}
\end{figure*}

\begin{figure*}
    \centering
    \begin{subfigure}{\textwidth}
    \centering
        \includegraphics[height=3.1cm]{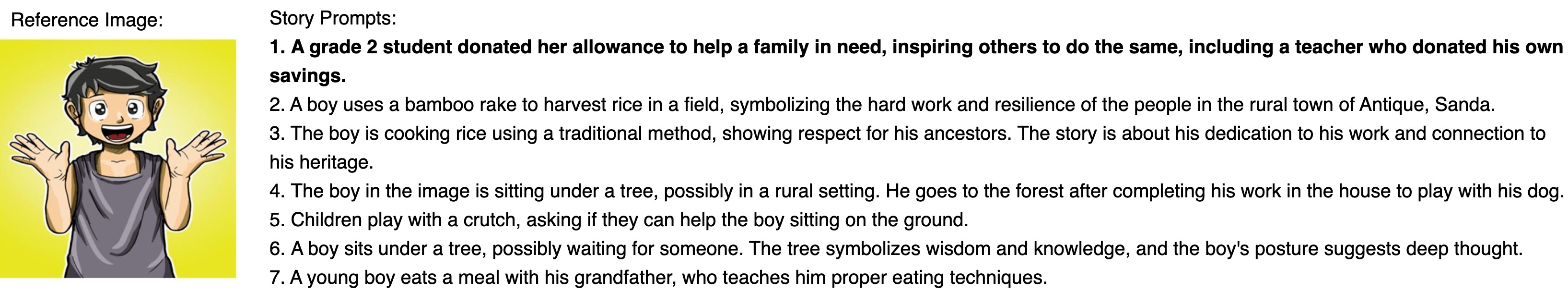}
    \end{subfigure}
    \begin{subfigure}{\textwidth}
        \raisebox{1.2cm}{\rotatebox[origin=c]{90}{Ground Truth}}
        \hfill
        \includegraphics[width=0.97\textwidth]{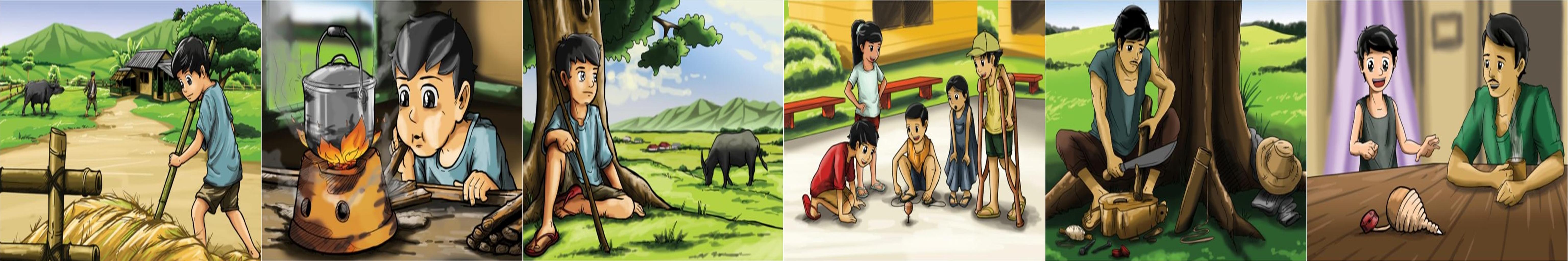}
    \end{subfigure}

    \begin{subfigure}{\textwidth}
        \raisebox{1.2cm}{\rotatebox[origin=c]{90}{SDXL-Prompt}}
        \hfill
        \includegraphics[width=0.97\textwidth]{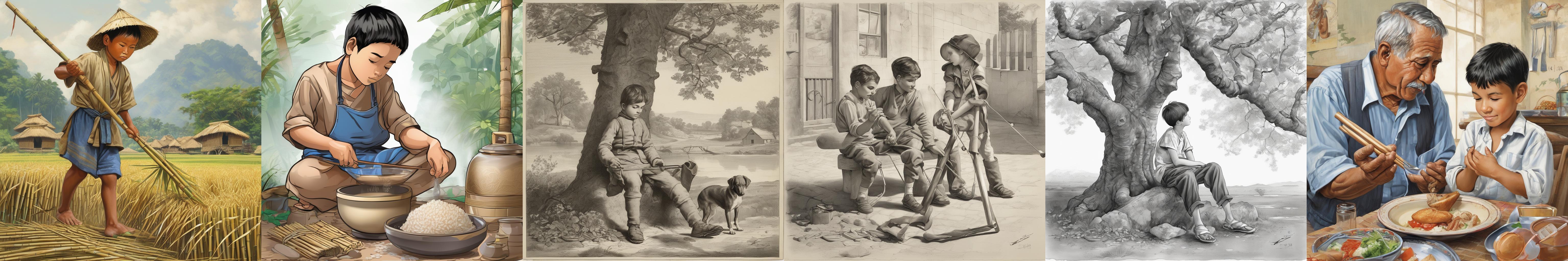}
    \end{subfigure}

    \begin{subfigure}{\textwidth}
        \raisebox{1.2cm}{\rotatebox[origin=c]{90}{IP-Adapter}}
        \hfill
        \includegraphics[width=0.97\textwidth]{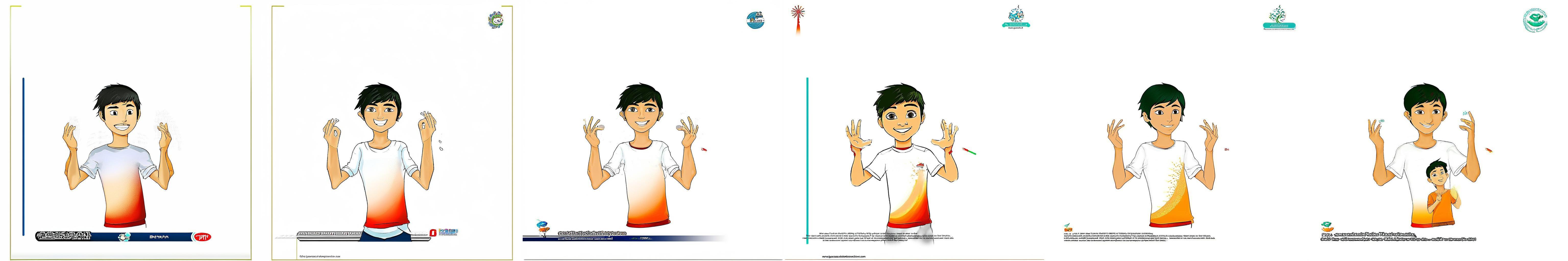}
    \end{subfigure}

    \begin{subfigure}{\textwidth}
        \raisebox{1.2cm}{\rotatebox[origin=c]{90}{StoryGen}}
        \hfill
        \includegraphics[width=0.97\textwidth]{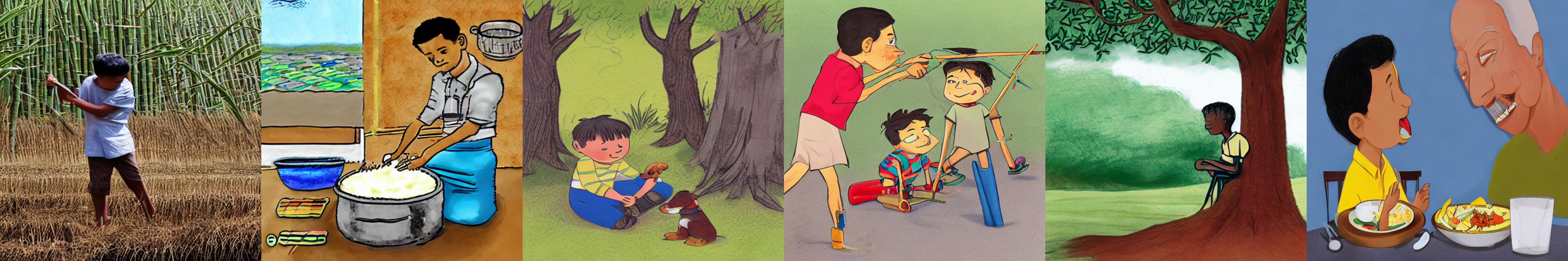}
    \end{subfigure}

    \begin{subfigure}{\textwidth}
        \raisebox{1.2cm}{\rotatebox[origin=c]{90}{Ours}}
        \hfill
        \includegraphics[width=0.97\textwidth]{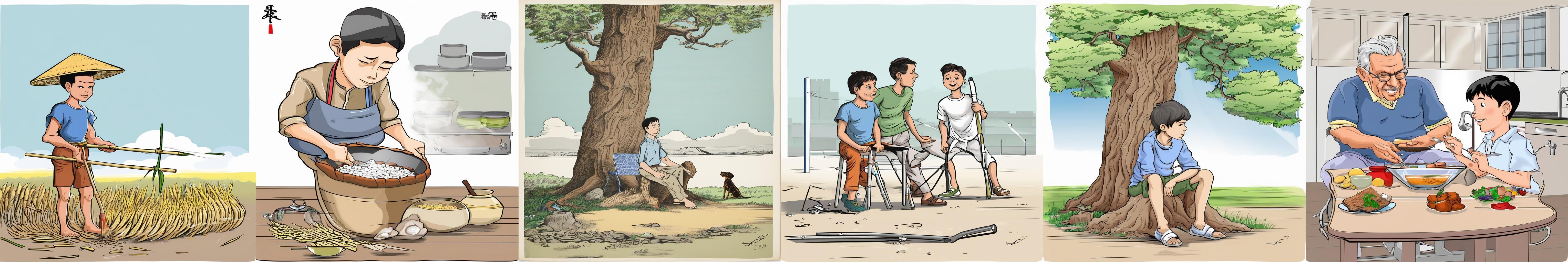}
    \end{subfigure}

    \caption{\textbf{Additional Qualitative results.} This figure presents a comparison of storytelling between ViSTA, baseline methods, and state-of-the-art on a sample StorySalon story. While SDXL-Prompt show high-quality and well-aligned images, they fail in generating consistent character acorss all frames. Although IP-Adapter shows consistent character, the generated images do not align with the prompt. Compare with the state-of-the-art StoryGen, our ViSTA shows better consistency on both characters and style.}
    \label{fig:x4}
\end{figure*}

Additional qualitative results from the StorySalon dataset are presented in Figures~\ref{fig:x1}, \ref{fig:x2}, \ref{fig:x3}, and \ref{fig:x4}. Consistent patterns are observed across different stories, including those featuring both human characters and animals, demonstrating that ViSTA achieves superior character and style consistency compared to other state-of-the-art baseline methods.

\subsection{FlintStonesSV Dataset}
\begin{figure}
    \centering
    \begin{subfigure}{\linewidth}
    \centering
        \includegraphics[width=0.95\linewidth]{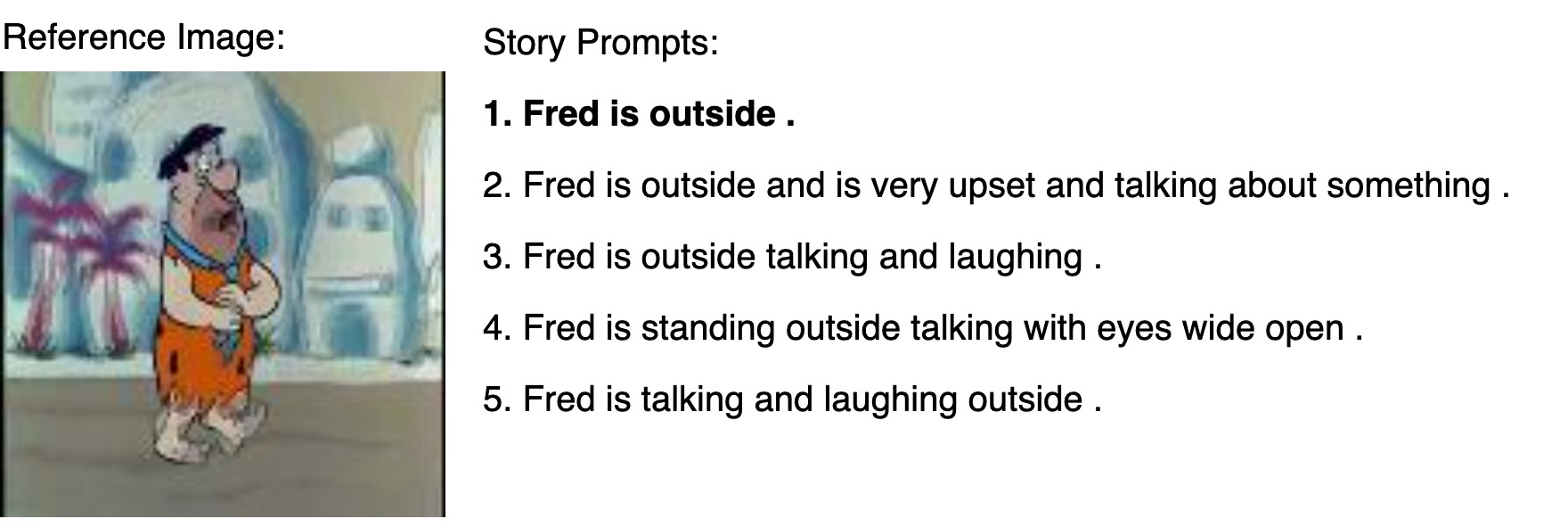}
    \end{subfigure}
    \begin{subfigure}{\linewidth}
    \centering
        \includegraphics[width=\linewidth]{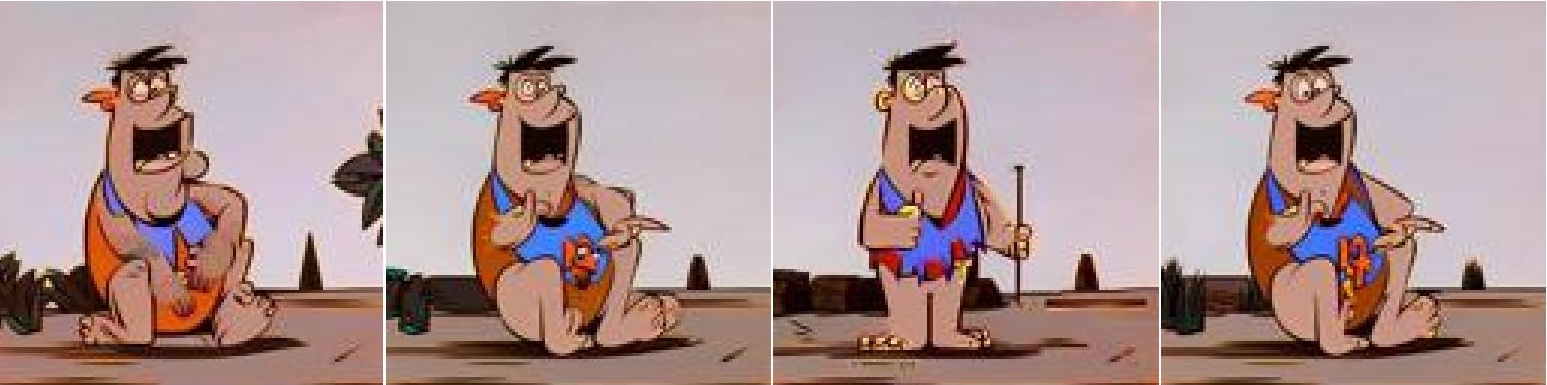}
    \end{subfigure}
    \caption{ViSTA sample on FlintStonesSV test set. }
    \label{fig:flintstone_x2}
\end{figure}

\begin{figure}
    \centering
    \begin{subfigure}{\linewidth}
    \centering
        \includegraphics[width=0.95\linewidth]{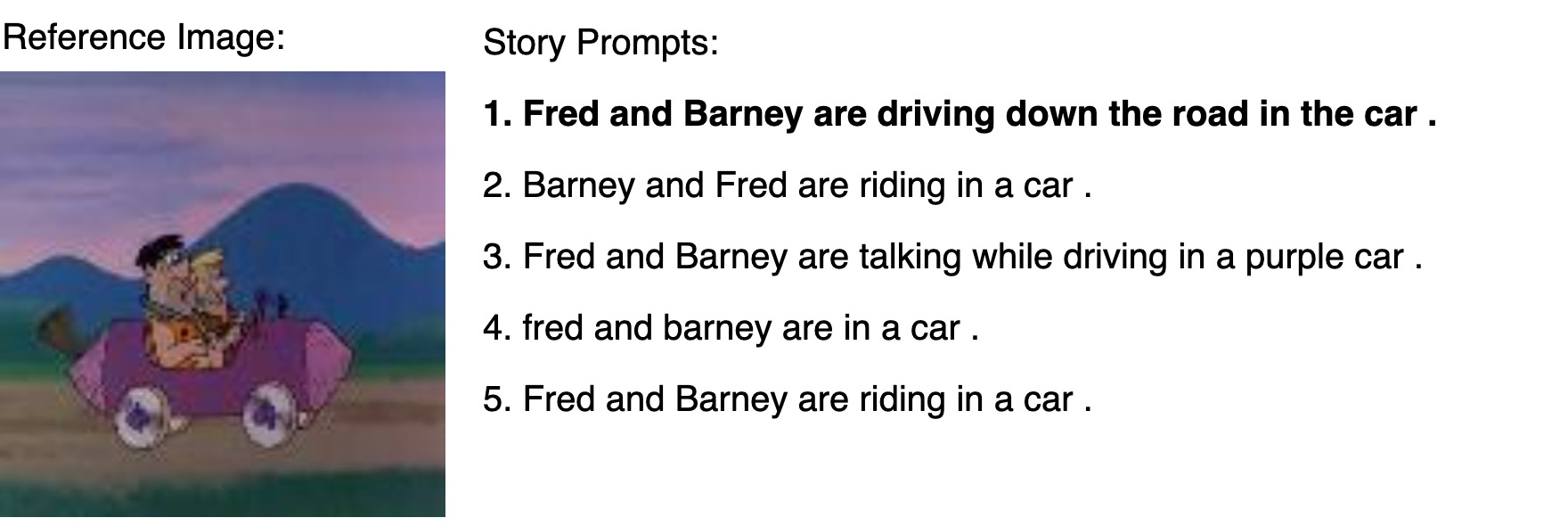}
    \end{subfigure}
    \begin{subfigure}{\linewidth}
    \centering
        \includegraphics[width=\linewidth]{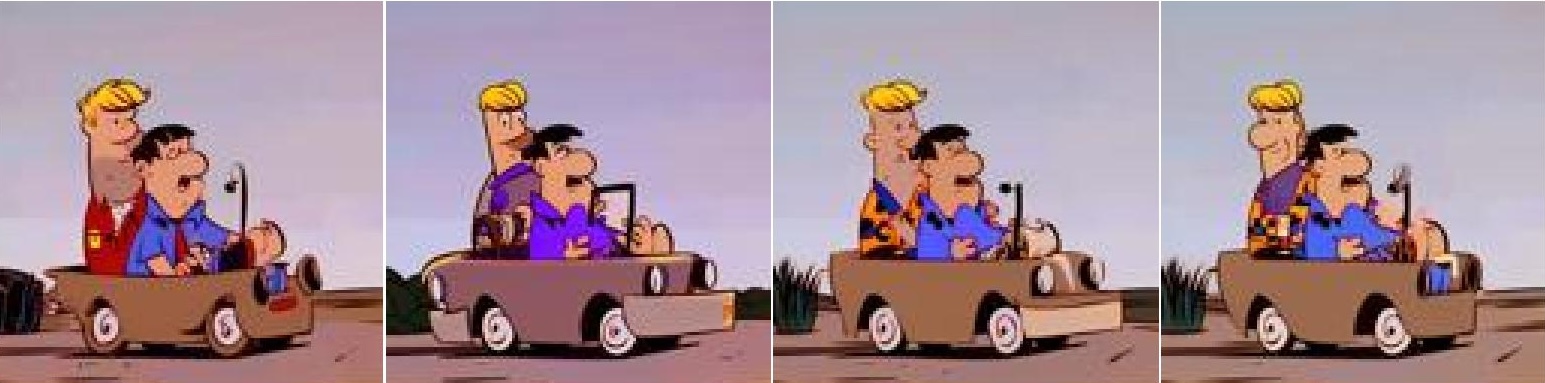}
    \end{subfigure}
    \caption{ViSTA sample on FlintStonesSV test set. }
    \label{fig:flintstone_x4}
\end{figure}

Figures~\ref{fig:flintstone_x2} and \ref{fig:flintstone_x4} illustrate examples generated by ViSTA from the FlintstonesSV dataset. The results demonstrate ViSTA's effectiveness in maintaining consistent character identity and coherent visual storytelling. 

\section{Human Evaluation}
To evaluate the quality of the generated story sequences, we conducted a pilot human study with five graduate student participants. Each participant was asked to assess the outputs from four different story generation methods.

The evaluation began with a set of detailed instructions provided through the survey interface, as shown below. These instructions outlined the evaluation procedure and defined the three core criteria: textual alignment, image alignment, and consistency.

\begin{tcolorbox}[colback=gray!10, colframe=gray!80, title=Story Generation Methods Assessment]
In this survey you will compare 4 different story generation methods that use state-of-the-art stable diffusion models across 5 examples. All methods start with a single prompt with its corresponding image as reference. After that, they generate a single image for each subsequent story prompt. For each example you will see the reference prompt-image pair and the subsequent story prompts, in addition to the original story book for reference on how the story looks like in a real-life example. You will be asked to compare the generated story based on 3 factors/metrics:
\begin{enumerate}
    \item Textual Alignment: how well does each frame follow its corresponding prompt
    \item Image Alignment: how well does each frame follow the overall style of the reference image
    \item Consistency: how well is the frame-to-frame consistency (same characters/objects)
\end{enumerate}
You will rank each metric separately for all methods giving 1 to the best method and 4 to the worst. Below are some examples of good/bad generations based on each metric.
\end{tcolorbox}

\begin{figure}[t]
    \centering
    \includegraphics[width=\linewidth]{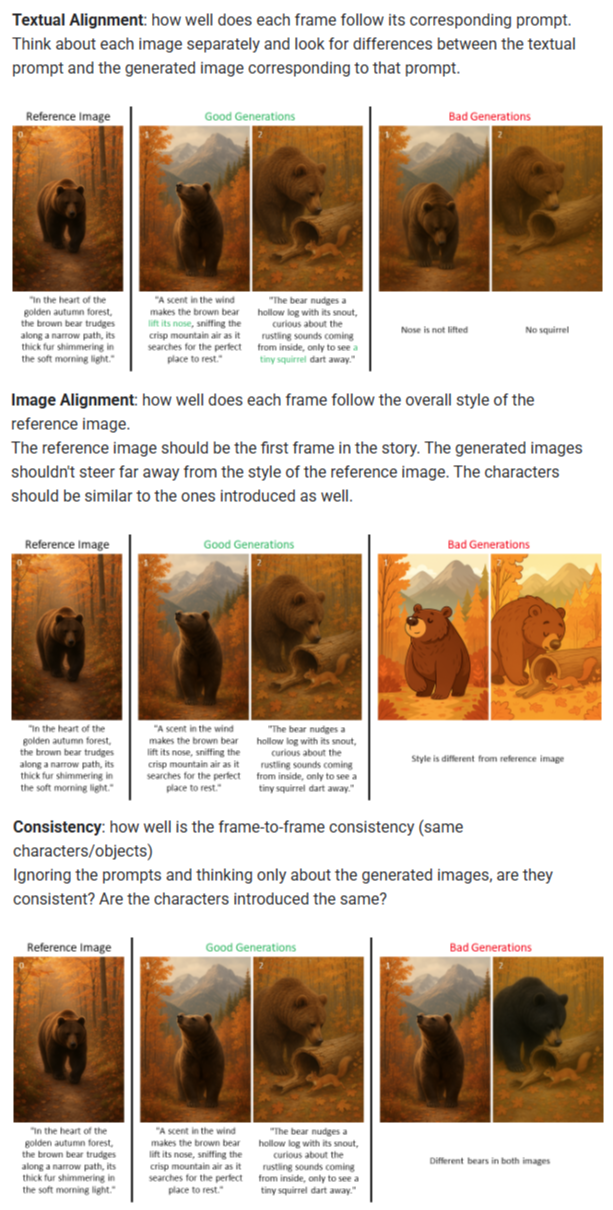}
    \caption{Examples illustrating good and poor outputs for each of the three evaluation metrics: textual alignment, image alignment, and consistency. These examples were shown to participants to calibrate their understanding of the evaluation criteria.}
    \label{fig:userstudy_eg}
\end{figure}

Figure~\ref{fig:userstudy_eg} presents the detailed explanations of each evaluation criterion, along with visual examples illustrating both high-quality and low-quality generations. These examples were included to calibrate participants' understanding and ensure consistent evaluation across annotators.

\begin{figure}[t]
    \centering
    \includegraphics[width=\linewidth]{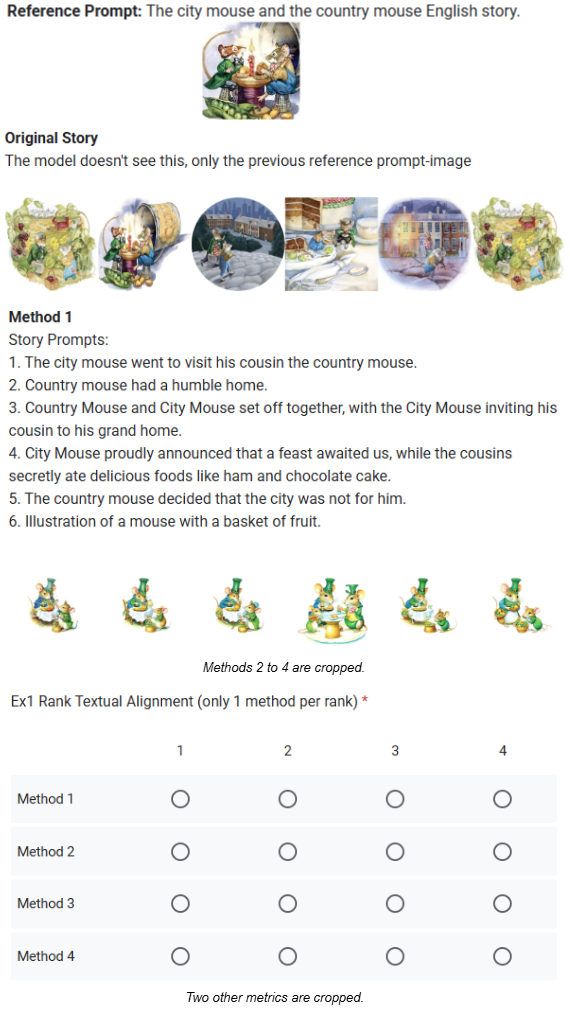}
    \caption{Interface for the human evaluation study. Each example included a reference image and prompt, original story content, and the outputs from four different generation methods (only showing one method for brevity). Participants ranked the methods for each metric independently (only showing one metric for brevity).}
    \label{fig:userstudy}
\end{figure}

Figure~\ref{fig:userstudy} illustrates the evaluation interface and procedure. For each of the five examples, participants were presented with a reference prompt and image, as well as the original story images. They then reviewed the image sequences generated by all four methods, shown in randomized order. For each criterion, participants independently ranked the four methods from 1 (best) to 4 (worst). 

Each participant evaluated five examples, resulting in a total of 25 sets of rankings per metric across the study.

\section{Limitations} 
Despite its effectiveness in improving character consistency and text-image alignment, our proposed ViSTA model has certain limitations. First, while our salient history selection strategy enhances efficiency by selecting the most relevant historical reference, it may still struggle in complex narratives where multiple past frames contribute equally to the current generation. 
The second limitation is the inherent limitation of auto-regressive methods, error accumulation. As minor imperfections in earlier frames propagate through the sequence, they can compound over time, leading to visual drift, degraded character consistency, or unintended distortions. While our salient history selection strategy mitigates this issue by prioritizing the most relevant historical references, it does not completely eliminate the risk of accumulated inconsistencies.
Lastly, our method is evaluated on animated datasets, and its generalizability to photorealistic or highly diverse artistic styles remains an open question.



